\documentclass{article}

\usepackage[preprint]{neurips_2026}

\usepackage[utf8]{inputenc} % allow utf-8 input
\usepackage[T1]{fontenc}    % use 8-bit T1 fonts
\usepackage{hyperref}       % hyperlinks
\usepackage{url}            % simple URL typesetting
\usepackage{booktabs}       % professional-quality tables
\usepackage{amsfonts}       % blackboard math symbols
\usepackage{nicefrac}       % compact symbols for 1/2, etc.
\usepackage{microtype}      % microtypography
\usepackage{xcolor}         % colors

\usepackage{amsmath}
\usepackage{graphicx}
\usepackage{tabularx}  
\usepackage{multirow} 
\usepackage{makecell}
\usepackage{wrapfig}
\usepackage{pifont}
\usepackage{enumitem}
\usepackage{subcaption}
\usepackage{colortbl}

\newcommand{\mybench}{HLL}

\title{HLL: Can Agents Cross Humanity’s Last Line of Verification?}

\author{
  Xinhao Song$^{1}$ \quad
  Su Su$^{2}$ \quad
  Sirui Song$^{1}$ \quad
  Hongliang Wu$^{1}$ \quad
  Wen Shen$^{3}$ \\
  \textbf{Zhihua Wei}$^{3}$ \quad
  \textbf{Gongshen Liu}$^{1}$ \quad
  \textbf{Linfeng Zhang}$^{1}$\thanks{Corresponding authors.} \quad 
  \textbf{Dongrui Liu}$^{1}$\footnotemark[1] \\
  \\
  $^{1}$Shanghai Jiao Tong University \quad
  $^{2}$Shandong University \quad
  $^{3}$Tongji University \\
  \\
  \texttt{\{sxh001\}@sjtu.edu.cn} 
}

\begin{document}

\maketitle

\begin{abstract}
Multimodal agents are increasingly expected to operate interfaces on behalf of users, raising a central deployment question: can they truly substitute for humans in workflows that services deliberately protect against automation? CAPTCHA verification makes this question concrete. It is not merely a visual puzzle, but a human-verification boundary placed before account creation, content access, form submission, and other protected actions. We introduce \textbf{Humanity's Last Line of Verification (HLL)}, a controlled benchmark that uses interactive CAPTCHA verification to evaluate whether agents can cross this boundary through grounded, human-like interaction rather than recognition alone. HLL covers diverse CAPTCHA interactions and exposes agents to controlled realism stressors, including cluttered webpages, harder task variants, and trace-conditioned validation of the solving process. We evaluate eight frontier multimodal agents in a closed-loop GUI environment. 
The results show that current agents remain brittle at this human-substitution boundary: performance varies sharply across verification types, degrades under realistic interface conditions, and drops further when correct answers must be supported by valid action traces. By exposing gaps in localization, action calibration, state tracking, and process consistency, HLL provides a concrete testbed for measuring how close multimodal agents are to acting as human substitutes in protected real-world workflows.  Our code is available at \url{https://github.com/XinhaoS0101/HLL}.
\end{abstract}

\section{Introduction}
\label{sec:intro}

\begin{wrapfigure}{R}{0.48\textwidth}
    \centering
    \vspace{-15pt}
    \includegraphics[width=\linewidth]{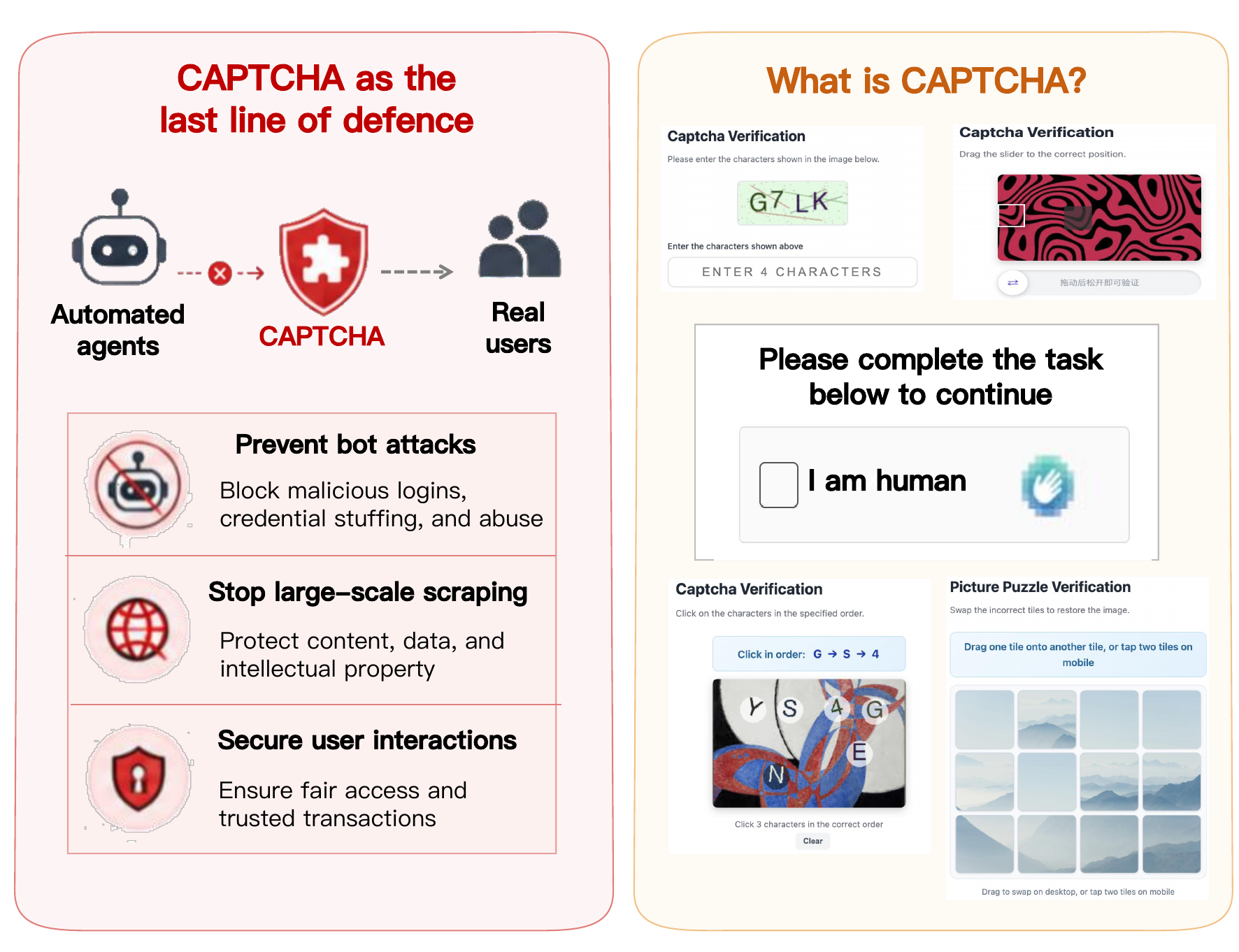}
    \captionof{figure}{CAPTCHA as the final frontier: securing web services by testing interactive, human-level reasoning against automated agents.}
    \label{fig:teaser1}
    \vspace{-10pt} 
\end{wrapfigure}

Multimodal agents are beginning to occupy a role that was previously reserved for human users: operating graphical interfaces to complete open-ended tasks. Systems such as OpenClaw and native GUI agents~\cite{qin2025ui,zhang2025appagent} point to agents that can act through browsers and mobile apps. However, completing user workflows in the wild requires more than navigation, screen understanding, or prediction of the final-answer. Agents must also cross the verification boundaries that real services place before account creation, content access, form submission, and transactions. This paper studies that human-substitution boundary: \textbf{whether general agents can pass the verification steps that real services use to decide who is allowed to proceed?}

CAPTCHAs make this boundary concrete. Without a human-verification step, automated programs can create fake accounts, scrape protected content, submit spam forms, abuse promotions, or repeatedly query services in ways that ordinary interface design cannot reliably prevent. For example, a website may insert a CAPTCHA before search, login, signup, or checkout when the traffic pattern resembles a crawler, so that access is conditioned on evidence of human presence rather than on the mere ability to send requests. CAPTCHAs were introduced for this purpose~\cite{CAPTCHA2003} and have since evolved from distorted text and image recognition~\cite{algwil2023survey,searles2023empirical} toward visual reasoning, clicking, dragging, and behavioral verification~\cite{gao2021vttsolver,ding2025illusioncaptcha}. For general-purpose agents, CAPTCHA solving is therefore not a niche recognition problem. It is a last-mile verification barrier and a compact stress test for whether general agents can pass a defense explicitly designed to separate humans from bots.

\begin{wrapfigure}{R}{0.48\textwidth}
    \centering
    \vspace{-15pt}
    \includegraphics[width=\linewidth]{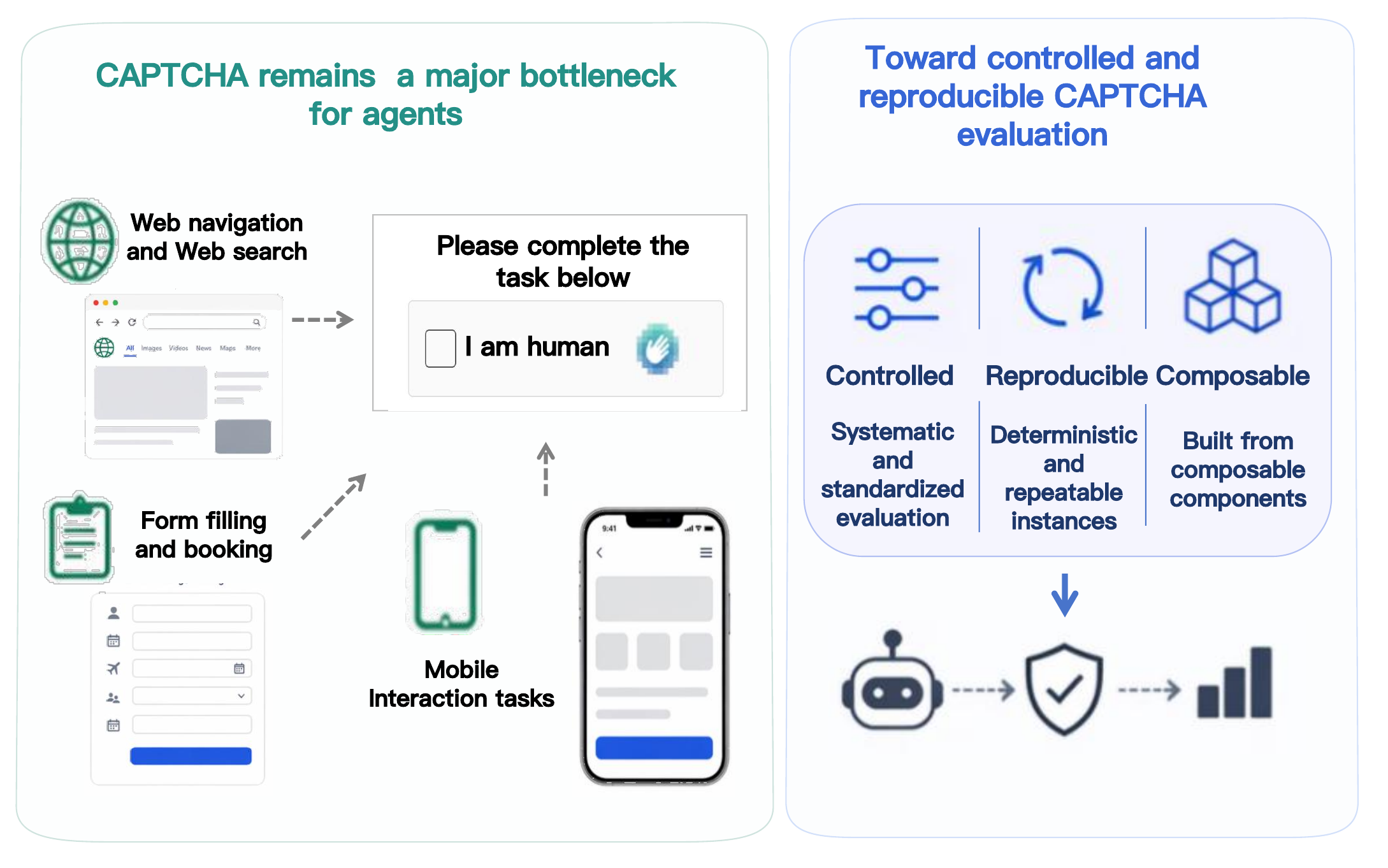}
    \captionof{figure}{Limitations of existing benchmarks: current static tasks fail to capture the complex, grounded interactions required in realistic environments.}
    \label{fig:teaser2}
    \vspace{-14pt}
\end{wrapfigure}

This last-mile verification barrier is largely missing from current agent evaluation. Existing web and GUI benchmarks measure progress on navigation and application control~\cite{pan2024webcanvas,mind2web}, browsing and web-game capabilities~\cite{wei2025browsecomp,thomas2025webgames}, or general task completion~\cite{liu2023agentbench}, but verification steps are often filtered out or treated as incidental obstacles rather than central capabilities. CAPTCHA-oriented work has a complementary limitation: many evaluations have largely treated verification either as an incidental obstacle in broader web tasks~\cite{luo2025open} or as a CAPTCHA-specific recognition and pass-rate problem~\cite{wu2026mca,deng2025oedipus}. Neither setting fully captures the realistic interaction process. To assess whether agents are ready to act as human substitutes, verification should instead be evaluated as an end-to-end deployment bottleneck. Such an evaluation must test the full perception-action loop required to proceed through protected interfaces, expose agents to controlled realism factors such as task difficulty, surrounding webpage distraction, and interaction-dependent validation, and diagnose whether success is supported by a valid solving process rather than a lucky final response. Figure~\ref{fig:teaser2} illustrates why this matters: a visually compact challenge can require perception, localization, grounded action, state tracking, and final validation, and the same final answer can mask very different failure modes along this pipeline.

To address this gap, we introduce \textbf{Humanity's Last Line of Verification (HLL)}, a controlled benchmark for evaluating interactive verification as a core capability of general multimodal agents. \mybench \ spans ten CAPTCHA families with heterogeneous interaction requirements and organizes evaluation along three orthogonal realism axes: intrinsic task difficulty, webpage distraction, and dynamic interaction validation. In dynamic settings, the benchmark moves beyond semantic correctness by collecting structured \textit{telemetry} through a unified submission payload and applying family-specific rules that test whether the final answer is supported by sufficient interaction evidence. This factorized design turns CAPTCHA solving from a single success flag into a diagnostic probe: it allows controlled comparisons between clean and cluttered environments, static and dynamic verification settings, and simpler versus harder task variants, revealing which component of the end-to-end interaction pipeline breaks first. Beyond task design, \mybench \ is implemented as a lightweight, agent-agnostic environment layer that can be integrated into both web-agent and mobile-agent evaluation pipelines with minimal adaptation.

The resulting benchmark is meant to make the web's last line of human verification measurable rather than anecdotal. By framing CAPTCHA solving as a factorized, end-to-end agent benchmark, \mybench \ provides a concrete testbed for studying how multimodal agents behave when perception, grounding, interface interaction, and anti-automation constraints are tightly coupled. In this sense, CAPTCHA solving becomes a lens for analyzing whether multimodal agents are ready to act as human substitutes in the wild.

In summary, this work makes the following core contributions:
\begin{itemize}[topsep=2pt, itemsep=2pt, parsep=0pt, leftmargin=20pt]
    \item \textbf{Formalization of Verification Bottlenecks:} We formalize interactive CAPTCHA solving as an end-to-end evaluation paradigm for testing the last-mile verification barrier faced by general agents acting on behalf of users.
    \item \textbf{Factorized Benchmark Design:} We introduce \mybench, featuring ten task families, a protocol that factorizes realism along three orthogonal axes: difficulty, distraction, and dynamic interaction validation, and a lightweight integration layer that supports both web-agent and mobile-agent evaluation.
    \item \textbf{Empirical Analysis and Failure Taxonomy:} We provide a systematic empirical study and a fine-grained failure taxonomy, showing that frontier agents suffer systematic degradation under realistic interaction and consistency constraints.
\end{itemize}
\vspace{-5pt}

\section{Related Work}
\label{sec:related-work}

\subsection{Multimodal and GUI Agents}
Recent multimodal agents increasingly operate in realistic interfaces, from web environments such as WebArena, VisualWebArena, Mind2Web, and WebCanvas~\cite{zhou2023webarena,koh2024visualwebarena,mind2web} to desktop and mobile settings such as OSWorld and AppAgent~\cite{xie2024osworld,zhang2025appagent}. Native GUI agents further improve screen perception and action generation~\cite{hong2024cogagent,qin2025ui,cheng2025kairos}. Existing benchmarks mainly evaluate general navigation, application control, or task completion~\cite{liu2023agentbench,pan2024webcanvas,wei2025browsecomp,thomas2025webgames}, while CAPTCHA-protected pages are often filtered out or treated as incidental obstacles. As a result, they provide limited evidence about whether agents can handle verification bottlenecks that require fine-grained perception, spatial grounding, action execution, and final submission under anti-automation constraints.

\subsection{CAPTCHA Benchmarks and Automated Solving}
CAPTCHAs were introduced as human verification mechanisms based on problems that were expected to be easy for humans but difficult for automated programs~\cite{CAPTCHA2003}. Early work studied text-based CAPTCHA usability and security~\cite{bursztein2011text,algwil2023survey,searles2023empirical}, while later attacks and defenses extended to image recognition~\cite{chellapilla2005building,sivakorn2016robot,noury2020deep}, visual reasoning~\cite{gao2021vttsolver,wang2018captcha,xue2025illusion}, adversarial examples~\cite{shi2021adversarial,hitaj2020capture}, and deployed reCAPTCHA-style systems~\cite{plesner2024breaking}. Recent benchmarks further broaden CAPTCHA evaluation: Open CaptchaWorld tests multimodal agents on diverse web-based CAPTCHA puzzles~\cite{luo2025open}, MCA-Bench studies multimodal CAPTCHA robustness~\cite{wu2026mca}, and LLM-enhanced solvers highlight the importance of spatial reasoning and multi-step inference~\cite{deng2025oedipus,teoh2025bot}. However, these works still largely emphasize recognition accuracy, attack success, or final-answer pass rates, which can obscure whether failures arise from perception, grounding, action execution, or process inconsistency.

\subsection{Interactive Verification as Perception-Action Evaluation}
Interactive CAPTCHA solving differs from static multimodal understanding benchmarks, where models often produce textual answers from fixed images, including VQA and text-aware tasks~\cite{goyal2017making,zhou2025egotextvqa} as well as broader multimodal reasoning suites~\cite{lu2023mathvista,yue2024mmmu,liu2024mmbench,yu2023mm}. In a browser loop, an agent must locate the verification region, ground actions to interface elements, adapt to intermediate state changes, and submit an answer supported by the solving process. Existing CAPTCHA benchmarks have begun to consider reasoning depth, action traces, or grounding annotations~\cite{luo2025open,deng2025oedipus,wu2026mca}, while visual-spatial reasoning benchmarks expose related reasoning gaps~\cite{liu2023visual,chollet2019measure,johnson2017clevr,ma20253dsrbench}. However, they do not fully factorize the realism conditions that make web verification difficult in practice. \mybench \ therefore evaluates heterogeneous CAPTCHA families under controlled axes of intrinsic difficulty, webpage distraction, and dynamic interaction validation.

\subsection{Robustness and Process-Level Validation}
A parallel line of work studies GUI-agent robustness under cluttered, adversarial, or dynamically changing environments. Agents can be misled by visual distractions and environmental injection~\cite{ma2024caution,chen2025evaluating,liao2024eia}, by pop-ups or adversarial web interfaces~\cite{zhang2025attacking,xu2024advagent}, and by broader multimodal or cross-modal attacks~\cite{wu2024dissecting,bagdasaryan2023abusing}. CAPTCHA systems are also process-oriented: many verification mechanisms rely on behavioral signals~\cite{acien2020becaptcha,acien2021becaptcha}, mouse or touch dynamics~\cite{khan2024mouse,frank2012touchalytics}, and continuous authentication signals~\cite{abuhamad2020sensor,zaidi2021touch} rather than final answers alone. \mybench \ does not reproduce full production-side bot detection, but introduces trace-conditioned dynamic validation to test whether an agent's answer is supported by task-consistent interaction evidence.
\vspace{-5pt}
\section{Benchmark}
\label{sec:benchmark}

In this section, we introduce \mybench \ as a controlled benchmark for measuring verification bottlenecks in multimodal agents. Rather than treating CAPTCHA solving as a narrow recognition problem, \mybench \ formulates it as an end-to-end interaction task in which an agent must perceive the challenge, localize the correct verification region, execute grounded interface actions, and satisfy anti-automation constraints under progressively more realistic web conditions. To make these bottlenecks measurable, \mybench \ organizes evaluation along factorized realism dimensions and supports fine-grained analysis of where the interaction pipeline breaks down. We first describe the benchmark design goals and formalize the benchmark unit, and then present the task families, realism axes, and evaluation protocol.

\begin{figure*}[t]
\centering
\includegraphics[width=\textwidth]{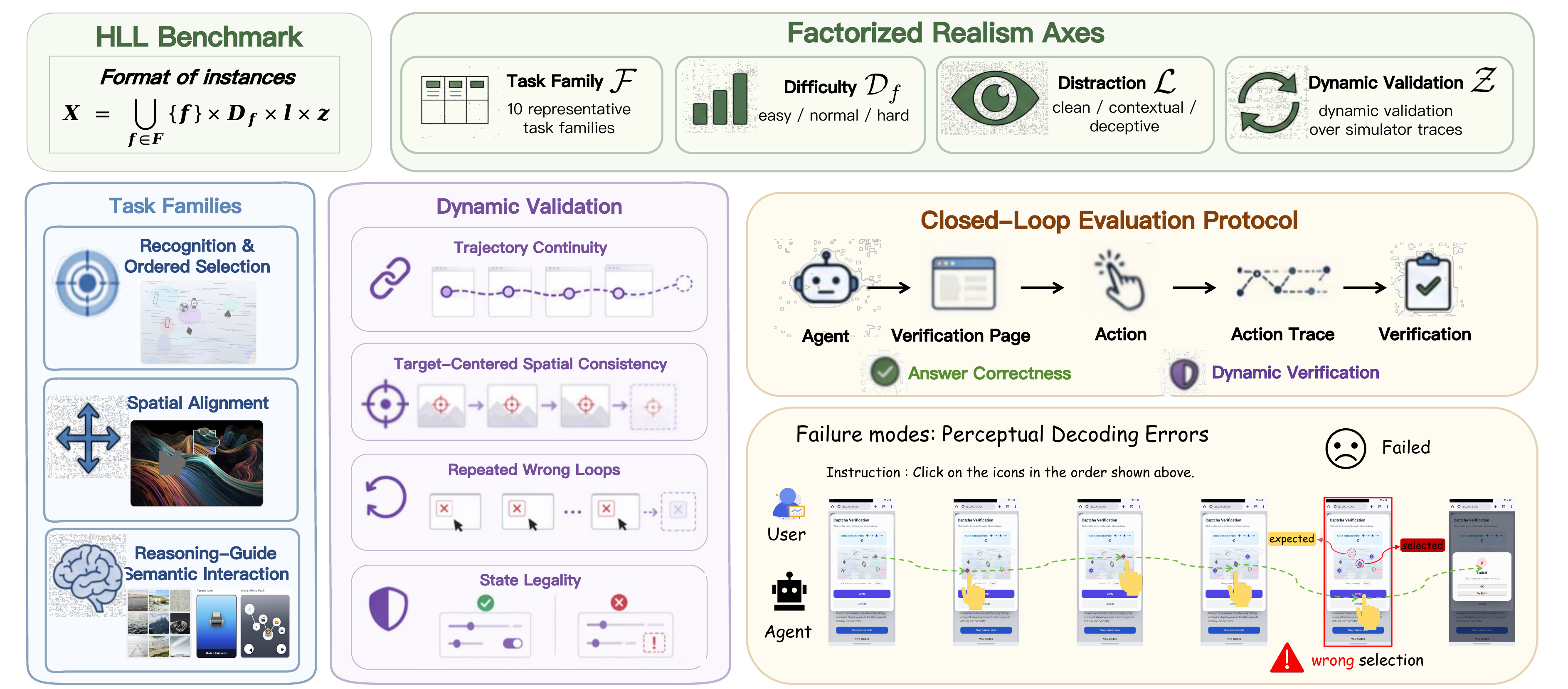}
\caption{Overview of the \mybench \ benchmark structure. The benchmark combines heterogeneous CAPTCHA task families with controlled realism axes, including intrinsic task difficulty, environmental distraction, and dynamic interaction validation, to expose where multimodal agents fail along the end-to-end verification pipeline.}
\label{fig:benchmark-overview}
\vspace{-12pt}
\end{figure*}

\subsection{Design Goals and Benchmark Formulation}
\label{subsec:design-goals}

The design of \mybench \ follows five goals: evaluate end-to-end verification beyond semantic recognition, cover heterogeneous interaction patterns, factorize realism into controllable dimensions, remain reproducible for closed-loop agents, and support interpretable failure analysis beyond a single success flag.

Following these design goals, we define the benchmark over a family set \(\mathcal{F}\), a family-dependent difficulty domain \(\mathcal{D}_f\), a distraction-level set \(\mathcal{L}\), a dynamic-validation flag \(z \in \{0,1\}\), and a sample index set \(\mathcal{S}\). One benchmark instance is a factorized tuple
\[
x = (f,d,\ell,z,s)
\quad \in \quad
\mathcal{X}
=
\bigcup_{f \in \mathcal{F}}
\{f\} \times \mathcal{D}_f \times \mathcal{L} \times \{0,1\} \times \mathcal{S}.
\]

Here, \(f\) specifies the semantic type of CAPTCHA challenge, \(d\) controls the intrinsic hardness of the task when supported, \(\ell\) determines how much irrelevant or deceptive webpage content is introduced around the CAPTCHA, \(z\) indicates whether the task additionally enforces interaction-consistency validation, and \(s\) indexes repeated instances under the same configuration. This formulation separates the core challenge type from the realism conditions under which it is evaluated.

We further associate each family with a dominant capability group through a mapping
\[
g:\mathcal{F}\rightarrow\mathcal{G},
\qquad
\mathcal{G}
=
\{\mathrm{Recognition},\mathrm{Spatial},\mathrm{Stateful},\mathrm{Reasoning}\},
\]
where \(g(f)\) is used for coarse-grained analysis rather than as a claim that the required capabilities are disjoint.

As summarized in Figure~\ref{fig:benchmark-overview}, this factorization enables controlled comparisons across clean and cluttered pages, static and dynamic validation, and easy versus hard variants. It shifts evaluation from whether a model solves a CAPTCHA to where and under which realism conditions the interaction pipeline fails.

\begin{wrapfigure}{R}{0.45\textwidth}
    \vspace{-20pt}
    \centering
    \includegraphics[width=0.95\linewidth]{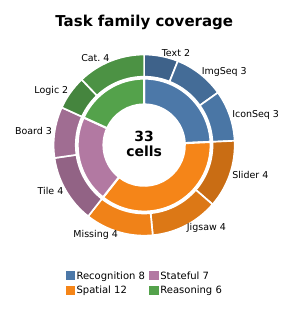}
    \vspace{-5pt} 
    \caption{Distribution of benchmark instances across CAPTCHA task families and realism-axis settings.}
    \label{fig:benchmark-distribution}
    \vspace{-35pt}
\end{wrapfigure}

Figure~\ref{fig:benchmark-distribution} further summarizes the benchmark composition across task families and realism-axis settings. 

\subsection{Task Families and Capability Taxonomy}
\label{subsec:task-families}

The semantic backbone of \mybench \ consists of ten \emph{base task families}. These ten families define the benchmark itself; later realism dimensions such as difficulty, distraction, and dynamic validation are layered on top of them and do not constitute new task families. 

At a coarse level, the benchmark covers four capability groups. \emph{Recognition and ordered selection} tasks test whether an agent can correctly perceive targets and interact with them in sequence, including text transcription, icon sequence selection, and natural-image sequence selection. \emph{Spatial alignment and local patch reasoning} tasks emphasize geometric precision and low-level action fidelity, including slider alignment, jigsaw alignment, and missing-patch selection. \emph{Stateful puzzle restoration} tasks require the agent to reason over intermediate state changes, including board reconfiguration and tile restoration. Finally, \emph{reasoning-guided semantic interaction} tasks stress structured interpretation and coordinated selection, including logic-and-arithmetic interaction and category-guided image selection.

Many task families involve overlapping mixtures of perception, grounding, reasoning, and execution, and \mybench \ does not claim a perfectly disjoint decomposition of capabilities. Instead, the grouping identifies the \emph{dominant} bottleneck each family is most likely to expose. Detailed task descriptions for all ten base families are deferred to Appendix~\ref{app:base-families}.

\subsection{Realism Axes and Dynamic Interaction Validation}
\label{subsec:realism-axes}

Beyond task-family coverage, \mybench \ is designed to evaluate agents under progressively more realistic verification conditions. To this end, the benchmark factorizes realism along three axes: intrinsic task difficulty, environmental distraction, and dynamic interaction validation. For an instance \(x=(f,d,\ell,z,s)\), we write its realism configuration as
\[
\rho(x) = (d,\ell,z),
\]
where \(d\) varies task-internal complexity, \(\ell\) varies environmental clutter, and \(z\) switches trace-conditioned validation on or off. These axes are layered on top of the same base task families so that the benchmark can compare agent behavior under controlled changes in realism rather than across entirely different tasks.

At the level of benchmark composition, each family contributes a set of supported evaluation cells. Let \(\mathcal{F}_{\mathrm{hard}}\) denote families with harder variants and \(\mathcal{F}_{\mathrm{dyn}}\) denote families with dynamic validation. The number of evaluation cells contributed by family \(f\) is
\[
m(f)
=
2
+
\mathbf{1}[f \in \mathcal{F}_{\mathrm{hard}}]
+
\mathbf{1}[f \in \mathcal{F}_{\mathrm{dyn}}],
\qquad
M=\sum_{f\in\mathcal{F}}m(f)=33,
\]
where the two base cells correspond to the clean static and distracted static settings.

\paragraph{Intrinsic Task Difficulty.}
For task families that naturally admit hardness control, \mybench \ varies the internal complexity of the challenge while preserving its semantic type. This allows us to ask whether a failure is caused by the core recognition-and-action problem itself or only emerges once ambiguity and precision requirements increase.

\paragraph{Environmental Distraction.}
\mybench \ also varies the amount of irrelevant or deceptive webpage content surrounding the true verification region. This axis is intended to capture a realistic gap between clean benchmark interfaces and real deployment settings, where an agent must first localize the relevant interaction region and avoid being diverted by unrelated controls or decoys.

\paragraph{Dynamic Interaction Validation.}
Finally, \mybench \ distinguishes between static correctness and dynamic interaction-grounded correctness. In static settings, success depends only on the final submitted answer. In dynamic settings, an answer is accepted only if it is accompanied by interaction evidence consistent with the intended solving process. The goal is to verify that the submitted answer is supported by a valid sequence of task-relevant interface actions.

Viewed together, these mechanisms provide interpretable anti-automation signals rather than a single opaque score. They capture whether the agent actually interacted with the page, whether a drag or click pattern looks behaviorally plausible, whether the agent falls into repeated erroneous loops, and whether state changes remain legally grounded under task rules. 
Detailed per-family rules, mechanism support, and failure categories are deferred to Appendix~\ref{app:difficulty}, Appendix~\ref{app:distraction}, and Appendix~\ref{app:dynamic}.

\subsection{Evaluation Protocol}
\label{subsec:eval-protocol}

\begin{table*}[t]
\centering
\renewcommand{\arraystretch}{1.15} 
\setlength{\tabcolsep}{4pt} 
\caption{Static benchmark results on the ten base task families under clean webpages and default task configurations. \textbf{Abbreviations:} SA (Slider Align.), ImS (Image Seq.), JA (Jigsaw Align.), LA (Logic-Arith.), CS (Category Select.), MP (Missing Patch), IcS (Icon Seq.), TR (Tile Rest.), BR (Board Reconf.). Entries report instance-level success rates (\%). \textbf{Bold} indicates the best performance, and \underline{underline} indicates the second best.}
\label{tab:static-base-results}
\resizebox{\textwidth}{!}{
\begin{tabular}{l cccccccccc >{\columncolor{red!10}}c}
\toprule
\textbf{Model} & 
\textbf{Text} & 
\textbf{SA} & 
\textbf{ImS} & 
\textbf{JA} & 
\textbf{LA} & 
\textbf{CS} & 
\textbf{MP} & 
\textbf{IcS} & 
\textbf{TR} & 
\textbf{BR} & 
\textbf{Avg.} \\
\midrule
\rowcolor{gray!15} \multicolumn{12}{c}{Leading Frontier Models} \\
GPT-5.4           & \textbf{100.00} & 68.50 & 43.50 & \underline{96.25} & 73.75 & \textbf{98.00} & \underline{96.75} & 25.25 & 36.50 & 61.50 & 70.00 \\
Gemini-3.1-Pro    & \textbf{100.00} & \underline{92.25} & \underline{85.75} & 76.50 & 67.50 & \underline{72.75} & 73.25 & 28.25 & \underline{53.75} & \textbf{88.00} & \underline{73.80} \\
Claude-Sonnet-4.6 & \textbf{100.00} & 48.25 & 23.75 & 36.50 & 17.50 & 62.75 & 15.25 & 22.25 & 0.00  & 27.75 & 35.40 \\
Claude-Opus-4.6   & \textbf{100.00} & \textbf{98.25} & \textbf{96.50} & \textbf{97.75} & \textbf{96.25} & \textbf{98.00} & \textbf{98.25} & \textbf{55.50} & \textbf{96.50} & \underline{63.00} & \textbf{90.00} \\
Grok-4            & \underline{97.75} & 34.25 & 78.50 & 87.50 & \underline{76.25} & 45.75 & 52.25 & \underline{32.50} & 15.25 & 62.00 & 58.20 \\

\rowcolor{gray!15} \multicolumn{12}{c}{Other Evaluated Models} \\
GLM-5V            & 42.25 & 75.50 & 0.00  & 2.50  & 37.50 & 0.00  & 0.00  & 4.25  & 0.00  & 0.00  & 16.20 \\
MiniMax-M2.7      & 82.50 & 11.50 & 16.25 & 47.75 & 24.25 & 1.75  & 18.00 & 0.00  & 0.00  & 0.00  & 20.20 \\
Qwen-Max          & 55.75 & 4.25  & 0.00  & 0.00  & 32.50 & 0.00  & 1.50  & 0.00  & 0.00  & 0.00  & 9.40  \\
\bottomrule
\end{tabular}
}
\vspace{-10pt}
\end{table*}
\mybench \ evaluates agents in a closed-loop interactive setting. For each instance \(x\), a policy \(\pi\) observes the page and emits GUI actions until submission, budget exhaustion, or timeout. We represent one rollout as
\[
\tau_{\pi}(x)
=
(o_0,a_0,o_1,a_1,\ldots,o_T,a_T),
\qquad
a_t \sim \pi(\cdot \mid o_{\leq t},a_{<t}),
\]
where \(o_t\) is the simulator observation and \(a_t\) is the grounded GUI action. This preserves the end-to-end nature of CAPTCHA solving: agents must infer, localize, manipulate, and submit under the designated realism conditions.

Let \(\hat{y}(\tau)\) be the final submitted answer, \(y_x\) the ground truth, and \(V_f(\tau)\in\{0,1\}\) the family-specific dynamic validator. Static success is
\[
S_{\mathrm{static}}(x,\tau)
=
\mathbf{1}[\hat{y}(\tau)=y_x],
\]
while dynamic success additionally requires trace-conditioned validation:
\[
S(x,\tau)
=
S_{\mathrm{static}}(x,\tau)
\cdot
\begin{cases}
1, & z=0,\\
V_f(\tau), & z=1.
\end{cases}
\]
Thus, dynamic evaluation adds simulator-observable interaction evidence while remaining agnostic to the model's internal reasoning.
\vspace{-5pt}

\section{Experiments}
\label{sec:experiments}

This section uses \mybench \ to study how current multimodal agents behave under interactive verification workloads. Our evaluation is designed to answer the following research questions:

\textbf{RQ1} How well do frontier agents solve the base static CAPTCHA tasks, and which capability gaps persist even without additional realism stress?

\textbf{RQ2} How does performance change when controlled realism factors---webpage distraction and increased intrinsic difficulty---are introduced?

\textbf{RQ3} To what extent does trace-conditioned dynamic validation further degrade agent success beyond static correctness?

\textbf{RQ4} What failure patterns recur across task families, and how do they relate to the capability dimensions exposed by the benchmark?

\begin{wrapfigure}{R}{0.45\textwidth}
    \vspace{-20pt}
    \centering
    \includegraphics[width=\linewidth]{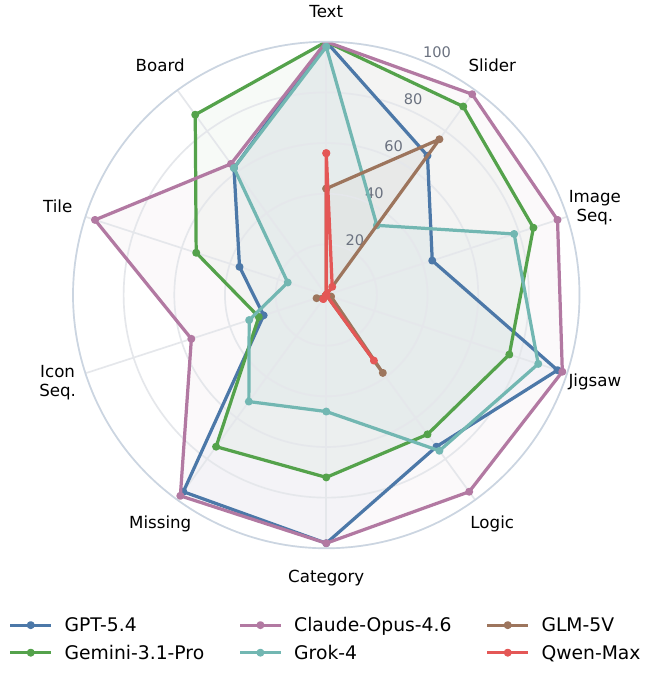}
    \caption{Static performance across the ten CAPTCHA task families.}
    \label{fig:static-radar}
    \vspace{-20pt} 
\end{wrapfigure}

\subsection{Experimental Setup}
\label{subsec:exp-setup}
\paragraph{Evaluated Agents.}
We evaluate a diverse set of frontier multimodal agents spanning multiple model families. The proprietary systems include OpenAI's GPT-5.4~\cite{openai2026gpt54systemcard}, Google's Gemini-3.1-Pro~\cite{google2026gemini31pro}, Anthropic's Claude-Sonnet-4.6~\cite{anthropic2026claudesonnet46} and Claude-Opus-4.6~\cite{anthropic2025claudeopus46}, xAI's Grok-4~\cite{xai2025grok4modelcard}, GLM-5V~\cite{hong2026glm}, MiniMax-M2.7~\cite{minimax2026m27}, and Qwen-Max~\cite{Qwen-VL}. All models are deployed as closed-loop GUI agents in the \mybench \ evaluation environment: at each step, the agent observes a rendered verification page, issues interface actions, receives updated simulator observations, and continues interacting until it submits an answer, reaches the maximum interaction budget, or times out. Detailed model specifications and hyperparameter settings are provided in Appendix~\ref{app:exp-details}.

Experiments cover three settings: \emph{base static} tasks on clean webpages, \emph{realism stress} from webpage distraction or higher intrinsic difficulty, and \emph{dynamic} variants that additionally require trace-conditioned validation. These settings separate failures from base task understanding, realism stress, and interaction-grounded validation.

\paragraph{Metrics.}
The primary metric is instance-level success rate as defined in Section~\ref{subsec:eval-protocol}. In the static and realism-stress settings, success is determined by semantic correctness of the submitted answer. In the dynamic setting, success additionally requires passing all applicable dynamic validation checks. We report per-family and aggregate results for each setting, and provide detailed analyses and case studies in Appendix~\ref{app:exp-details}.

\subsection{Static Results on Base Tasks}
\label{subsec:static-base-results}

We begin with the base static setting, where agents are evaluated on the ten base task families under clean webpages and default task configurations. Table~\ref{tab:static-base-results} summarizes the corresponding results. Even in this simplified setting, interactive CAPTCHA solving remains uneven: the strongest model reaches high overall performance, but success still varies sharply across task families, and most agents remain far from robust across the full benchmark.

Three observations stand out. \textbf{First, the leading model is strong but not fully saturated}. Claude-Opus-4.6 obtains the highest average success rate, followed by Gemini-3.1-Pro and GPT-5.4. However, even Claude-Opus-4.6 does not solve all task families perfectly, with residual errors on ordered icon selection and board reconfiguration. This suggests that clean-page static success is already bottlenecked by nontrivial perception--action coordination rather than by simple visual recognition alone.

\begin{table*}[t]
\centering
\renewcommand{\arraystretch}{1.15} 
\setlength{\tabcolsep}{4pt} 
\caption{Results under webpage distraction. Entries report instance-level success rates (\%). \textbf{Bold} indicates the best performance, and \underline{underline} indicates the second best.}
\label{tab:distraction-results}
\resizebox{0.95\textwidth}{!}{
\begin{tabular}{l cccccccccc >{\columncolor{red!10}}c}
\toprule
\textbf{Model} & 
\textbf{Text} & 
\textbf{SA} & 
\textbf{ImS} & 
\textbf{JA} & 
\textbf{LA} & 
\textbf{CS} & 
\textbf{MP} & 
\textbf{IcS} & 
\textbf{TR} & 
\textbf{BR} & 
\textbf{Avg.} \\
\midrule
\rowcolor{gray!15} \multicolumn{12}{c}{Leading Frontier Models} \\
GPT-5.4           & \textbf{100.00} & \textbf{100.00} & 64.25 & \textbf{100.00} & 53.75 & \textbf{100.00} & \textbf{100.00} & \underline{22.25} & 37.75 & 26.00 & \underline{70.40} \\
Gemini-3.1-Pro    & \textbf{100.00} & \underline{85.75} & \textbf{92.25} & 46.25 & 67.75 & 56.50 & 58.00 & 13.50 & \underline{44.50} & \textbf{93.50} & 65.80 \\
Claude-Sonnet-4.6 & \textbf{100.00} & 28.25 & 23.75 & 12.50 & 15.50 & 46.25 & 17.75 & 0.00  & 2.50  & 13.50 & 26.00 \\
Claude-Opus-4.6   & \textbf{100.00} & 83.75 & \underline{84.25} & \underline{81.75} & \textbf{94.50} & \underline{85.75} & \underline{75.50} & \textbf{38.25} & \textbf{100.00} & \underline{48.25} & \textbf{79.20} \\
Grok-4            & \textbf{100.00} & 32.50 & 55.50 & 68.25 & 64.50 & 51.50 & 38.25 & 14.00 & 17.75 & 45.75 & 48.80 \\

\rowcolor{gray!15} \multicolumn{12}{c}{Other Evaluated Models} \\
GLM-5V            & \underline{77.75} & 0.00  & 2.25  & 0.00  & \underline{75.75} & 4.50  & 0.00  & 1.75  & 0.00  & 0.00  & 16.20 \\
MiniMax-M2.7      & 74.50 & 33.50 & 18.25 & 56.50 & 21.75 & 2.25  & 0.00  & 0.00  & 0.00  & 1.25  & 20.80 \\
Qwen-Max          & 73.75 & 0.00  & 0.00  & 2.25  & 8.50  & 1.50  & 0.00  & 0.00  & 0.00  & 0.00  & 8.60  \\
\bottomrule
\end{tabular}
}
\vspace{-20pt}
\end{table*}

\begin{wraptable}{R}{0.5\textwidth}
    \centering
    \caption{Results on hard variants. \textbf{Bold}: best, \underline{underline}: second best.}
    \label{tab:hard-results}
    \renewcommand{\arraystretch}{1.15} 
    \setlength{\tabcolsep}{3pt} 
    
    \resizebox{\linewidth}{!}{
    \begin{tabular}{l ccccc >{\columncolor{red!10}}c}
    \toprule
    \textbf{Model} & \textbf{SA} & \textbf{JA} & \textbf{CS} & \textbf{MP} & \textbf{TR} & \textbf{Avg.} \\
    \midrule
    \rowcolor{gray!15} \multicolumn{7}{c}{Leading Frontier Models} \\
    GPT-5.4           & \underline{40.25} & \underline{20.25} & 24.75 & \textbf{79.75} & \underline{20.00} & \underline{37.00} \\
    Gemini-3.1-Pro    & 0.00  & \underline{20.25} & \underline{34.75} & 45.00 & \underline{20.00} & 24.00 \\
    Claude-Sonnet-4.6 & 0.00  & 10.25 & 9.75  & 10.00 & 0.00  & 6.00  \\
    Claude-Opus-4.6   & \textbf{65.25} & \textbf{74.75} & \textbf{70.00} & \underline{60.00} & \textbf{40.00} & \textbf{62.00} \\
    Grok-4            & 10.25 & 14.75 & 20.00 & 0.00  & 0.00  & 9.00  \\

    \rowcolor{gray!15} \multicolumn{7}{c}{Other Evaluated Models} \\
    GLM-5V            & 20.00 & 0.00  & 0.00  & 0.00  & 0.00  & 4.00  \\
    MiniMax-M2.7      & 20.00 & \underline{20.00} & 0.00  & 10.00 & 0.00  & 10.00 \\
    Qwen-Max          & 0.00  & 0.00  & 0.00  & 0.00  & 0.00  & 0.00  \\
    \bottomrule
    \end{tabular}
    }
    \vspace{-10pt} 
\end{wraptable}

\textbf{Second, task difficulty is highly uneven across families.} Text transcription is the easiest family overall, with several models achieving near-perfect or perfect performance. By contrast, ordered icon selection and tile restoration remain much harder on average, even though individual models can perform well on them. These families require tighter coupling between recognition and control: the agent must not only identify the target correctly, but also maintain sequential consistency, geometric precision, or state-aware manipulation throughout the interaction.

\begin{wrapfigure}{R}{0.45\textwidth}
    \vspace{-20pt}
    \centering
    \includegraphics[width=\linewidth]{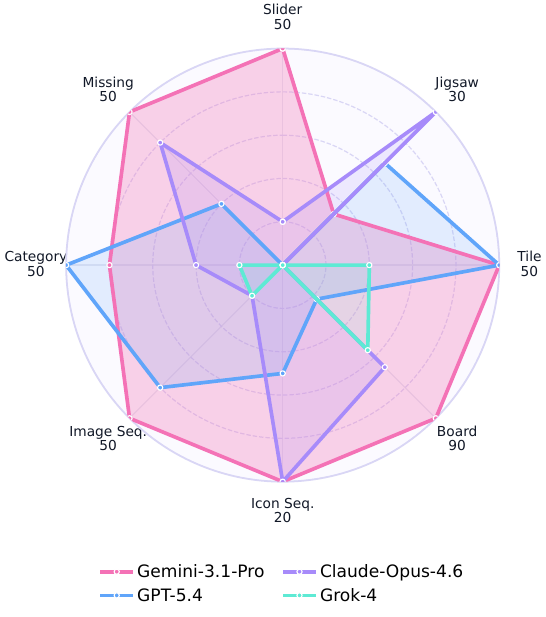}
    \caption{Dynamic performance across the CAPTCHA task families.}
    \label{fig:dynamic-radar}
    \vspace{-20pt} 
\end{wrapfigure}

\textbf{Third, aggregate averages conceal strong cross-family specialization.} The leading agents show complementary strengths across recognition, alignment, ordered selection, and stateful tasks, indicating that static CAPTCHA solving is not a single capability but a composition of partially separable skills. Figure~\ref{fig:static-radar} and Figure~\ref{fig:dynamic-radar} visualize these family-wise and cross-setting differences, with extended comparisons deferred to Appendix~\ref{app:full-static}.

\paragraph{Static failure patterns.}
The static traces show that failures are not reducible to visual recognition errors. Instead, they arise along the full interaction chain: weaker models misread distorted text or category cues, middle-performing models often identify the correct target but click or drag at imprecise coordinates, and stronger models mainly fail on ordered selection or stateful restoration tasks that require maintaining a structured interface state across multiple actions. Detailed trace-level examples are provided in Appendix~\ref{app:static-failure-cases}.

\vspace{-4pt}

\subsection{Robustness under Realism Stress}
\label{subsec:realism-stress-results}

We next evaluate how agent performance changes when the clean base setting is made more realistic. We consider two complementary stressors introduced in Section~\ref{subsec:realism-axes}: webpage distraction and increased intrinsic task difficulty. 

\textbf{Webpage distraction amplifies existing grounding and control weaknesses rather than introducing a uniform failure mode.} Table~\ref{tab:distraction-results} shows that most agents degrade under irrelevant or deceptive surrounding content, but the degradation is uneven. Claude-Opus-4.6 remains the strongest model overall, while GPT-5.4 and Gemini-3.1-Pro retain competitive average performance. In contrast, weaker models degrade to near-zero performance on many non-text families.
\vspace{-2pt}

\textbf{Increased intrinsic difficulty causes broad performance drops even for the strongest models.} Unlike distraction, which stresses page-level localization, difficulty stresses the task itself through tighter alignment tolerances, more ambiguous candidates, stronger visual confusion, or more complex restoration states. Table~\ref{tab:hard-results} reports results on the hard variants. Claude-Opus-4.6 remains the most robust under hard settings, while GPT-5.4 and Gemini-3.1-Pro fall sharply, especially on alignment and visual-restoration families. These results separate failures caused by environmental clutter from failures caused by the underlying CAPTCHA operation becoming intrinsically harder. Detailed stress analyses are provided in Appendix~\ref{app:full-distraction} and Appendix~\ref{app:full-difficulty}.

\vspace{-5pt}

\subsection{Dynamic Interaction Validation}
\label{subsec:dynamic-results}

We finally evaluate dynamic variants, where a semantically correct answer must additionally satisfy trace-conditioned validation rules over simulator-observable interaction traces and committed page states. Table~\ref{tab:dynamic-results} reports success rates on the eight task families with dynamic validation support. Compared with the static setting, the dynamic setting introduces a much stricter requirement: agents must not only infer the correct final answer, but also produce an interaction process that is consistent with the intended solving procedure.

Three observations stand out. \textbf{First, dynamic validation substantially lowers success even for models that perform well in the static setting.} Gemini-3.1-Pro achieves the best dynamic average at 45.0, while GPT-5.4 and Claude-Opus-4.6 drop to 26.3 and 23.8. 

\begin{table*}[t]
\centering
\renewcommand{\arraystretch}{1.15} 
\setlength{\tabcolsep}{4pt} 
\caption{Dynamic validation results. The \textbf{Static Avg.} column (blue) reports the matched static baseline averaged over the same eight dynamic-supported task families. Entries report instance-level success rates (\%). \textbf{Bold}: best, \underline{underline}: second best.}
\label{tab:dynamic-results}
\resizebox{0.95\textwidth}{!}{
\begin{tabular}{l cccccccc >{\columncolor{blue!10}}c >{\columncolor{red!10}}c}
\toprule
\textbf{Model} & 
\textbf{SA} & 
\textbf{JA} & 
\textbf{TR} & 
\textbf{BR} & 
\textbf{IcS} & 
\textbf{ImS} & 
\textbf{CS} & 
\textbf{MP} & 
\textbf{Static Avg.} & 
\textbf{Dyn. Avg.} \\
\midrule
\rowcolor{gray!15} \multicolumn{11}{c}{Leading Frontier Models} \\
GPT-5.4           & 0.00  & \underline{18.50} & \textbf{50.25} & 18.75 & \underline{11.25} & \underline{38.50} & \textbf{51.50} & 21.25 & 65.8 & \underline{26.3} \\
Gemini-3.1-Pro    & \textbf{48.25} & 12.50 & \textbf{49.75} & \textbf{90.25} & \textbf{19.75} & \textbf{50.50} & \underline{38.75} & \textbf{50.25} & \underline{71.3} & \textbf{45.0} \\
Claude-Sonnet-4.6 & 0.00  & 0.00  & 0.00  & 0.00  & 0.00  & 0.00  & 0.00  & \underline{40.00} & 29.6 & 5.0  \\
Claude-Opus-4.6   & \underline{10.00} & \textbf{29.75} & 0.00  & \underline{60.25} & \textbf{20.25} & 8.75  & 21.00 & \underline{40.00} & \textbf{88.0} & 23.8 \\
Grok-4            & 0.00  & 0.00  & \underline{18.50} & 48.75 & 0.00  & 11.25 & 11.50 & 0.00  & 51.0 & 11.3 \\

\rowcolor{gray!15} \multicolumn{11}{c}{Other Evaluated Models} \\
GLM-5V            & \underline{10.00} & 0.00  & 0.00  & 0.00  & 0.00  & 0.00  & 0.00  & 0.00  & 10.3 & 1.3  \\
MiniMax-M2.7      & 0.00  & 0.00  & 0.00  & 0.00  & 0.00  & 0.00  & 0.00  & 20.00 & 11.9 & 2.5  \\
Qwen-Max          & 0.00  & 0.00  & 0.00  & 0.00  & 0.00  & 20.00 & 0.00  & 0.00  & 0.7  & 2.5  \\
\bottomrule
\end{tabular}
}
\vspace{-15pt}
\end{table*}

\textbf{Second, the dynamic setting reshuffles model rankings.} Claude-Opus-4.6 is the strongest model in the base static and hard settings, but its dynamic performance is lower than that of Gemini-3.1-Pro and GPT-5.4. The largest gap appears on tasks where the final answer can be semantically inferred but the intermediate action sequence must still satisfy task-specific constraints. This suggests that dynamic validation probes a distinct process-level capability rather than simply increasing the perceptual difficulty of the challenge.

\textbf{Third, failures concentrate in interaction-heavy task families.} Ordered icon selection, slider alignment, and jigsaw alignment remain difficult under dynamic validation, reflecting the need for target-centered clicks or continuous drag traces. Board reconfiguration is an exception for Gemini-3.1-Pro and Claude-Opus-4.6, suggesting that some models can maintain valid state transitions in structured board-like tasks. Additional dynamic analyses are provided in Appendix~\ref{app:full-dynamic}.

\vspace{-5pt}

\section{Conclusion}
\vspace{-3pt}
This paper introduced \mybench, a controlled benchmark for evaluating whether multimodal agents can cross the human-verification boundaries that protect real web workflows from automation. By treating interactive CAPTCHA solving as an end-to-end verification bottleneck, \mybench \ separates semantic recognition from localization, grounded action execution, state tracking, and trace-consistent completion. Our experiments show that current frontier agents remain far from uniformly robust at this human-substitution boundary: even clean static tasks expose strong family-level specialization, realism stress amplifies failures in grounding and precision, and dynamic validation reveals cases where plausible final answers are not supported by valid interaction evidence. Although \mybench \ does not exhaustively cover the full diversity or service-specific policies of production verification systems, these findings suggest that agents are not yet ready to act as reliable human substitutes in protected real-world workflows without stronger spatial grounding, action calibration, state tracking, and recovery under realistic web conditions.

\newpage
\bibliographystyle{plain}
\bibliography{reference}

\newpage
\appendix

\section{Detailed Benchmark Specification}
\label{app:benchmark}

This appendix provides detailed benchmark information complementary to Section~\ref{sec:benchmark}. In particular, we document the ten base task families, the construction of distraction levels, the definition of difficulty settings, the scope and mechanism of dynamic interaction validation, and the detailed benchmark statistics used in the experimental study.

\subsection{Base Task Families}
\label{app:base-families}

The semantic backbone of \mybench \ consists of ten base task families. These families define the benchmark itself; realism dimensions such as difficulty, distraction, and dynamic validation are layered on top of them and are not counted as new tasks.

\paragraph{Recognition and Ordered Selection.}
This group includes \emph{text transcription}, \emph{icon sequence selection}, and \emph{natural-image sequence selection}. These tasks require the agent to first identify the correct targets and then execute the required sequence of actions in order. Their main purpose is to probe visual recognition, instruction grounding, and ordering fidelity.

\paragraph{Spatial Alignment and Local Patch Reasoning.}
This group includes \emph{slider alignment}, \emph{jigsaw alignment}, and \emph{missing-patch selection}. These tasks stress fine-grained localization, geometric precision, and the ability to transform high-level task understanding into accurate low-level manipulation.

\paragraph{Stateful Puzzle Restoration.}
This group includes \emph{board reconfiguration} and \emph{tile restoration}. Unlike single-step alignment tasks, these families require the agent to reason over intermediate state changes and progressively move the interface toward a valid target configuration.

\paragraph{Reasoning-Guided Semantic Interaction.}
This group includes \emph{logic-and-arithmetic interaction} and \emph{category-guided image selection}. These tasks place greater emphasis on structured interpretation, semantic reasoning, and coordinated multi-step interaction before submission.

\paragraph{Per-family Task Details.}
For completeness, the ten base families in the benchmark are:
\begin{itemize}
    \item \textbf{Text transcription:} read a short visual code and enter it correctly.
    \item \textbf{Natural-image sequence selection:} click a sequence of targets in the required order over cluttered visual content.
    \item \textbf{Jigsaw alignment:} drag a missing component to its correct geometric position.
    \item \textbf{Slider alignment:} continuously drag a control until a visual gap is correctly aligned.
    \item \textbf{Category-guided image selection:} select all grid images matching a semantic category.
    \item \textbf{Tile restoration:} restore an image by swapping misplaced tiles.
    \item \textbf{Logic-and-arithmetic interaction:} solve arithmetic, symbolic, or interface-mediated reasoning challenges and submit the correct result.
    \item \textbf{Missing-patch selection:} select the correct local patch that completes an image.
    \item \textbf{Board reconfiguration:} manipulate a compact board until a valid five-in-a-row state is formed.
    \item \textbf{Icon sequence selection:} click symbolic targets in the specified sequence.
\end{itemize}

\subsection{Distraction Layer}
\label{app:distraction}

To simulate more realistic webpage conditions, \mybench \ augments base CAPTCHA tasks with an environment-level distraction layer. This layer is designed to preserve the semantic task while changing the perceptual and interaction context around it.

\paragraph{Level 0: Clean Interface.}
The CAPTCHA is presented without additional webpage clutter. This setting is used as the default clean benchmark condition.

\paragraph{Level 1: Realistic Webpage Context.}
The CAPTCHA is embedded in a realistic webpage-style environment with background content and a fixed-size verification dialog. This setting increases the need for localization and contextual grounding while keeping the true interaction region stable.

\paragraph{Level 2: Deceptive Webpage Context.}
In addition to the background webpage, the environment introduces decoy controls and trap-like interaction elements outside the true CAPTCHA region. This setting is intended to probe whether agents can distinguish relevant verification controls from misleading interface content.

\subsection{Difficulty Settings}
\label{app:difficulty}

Difficulty is implemented as an intrinsic task-level control and is only enabled for base families whose hardness can be adjusted without changing the semantic nature of the task. In the current benchmark, explicit difficulty control is applied to a subset of families, primarily those involving alignment, candidate ambiguity, or visual confusion.

\paragraph{General Principle.}
Across families, \texttt{easy} reduces ambiguity or precision demand, \texttt{hard} increases intrinsic ambiguity or control difficulty, and \texttt{normal} serves as the default intermediate setting. However, the concrete implementation differs substantially by family. Rather than treating difficulty as a global scalar, \mybench \ instantiates it through family-specific controls that preserve the original task semantics.
\begin{itemize}
    \item \textbf{Easy.}
Easy settings reduce ambiguity or control precision requirements. Typical mechanisms include fewer distractor candidates, simpler layouts, or looser alignment constraints.
    \item \textbf{Normal.}
Normal serves as the default benchmark setting and is used whenever no family-specific difficulty control is enabled.
    \item \textbf{Hard.}
Hard settings increase intrinsic task complexity while preserving task semantics. Depending on the family, this may involve higher distractor similarity, more candidate options, stricter spatial tolerance, or more ambiguous visual restoration targets.
\end{itemize}

\paragraph{Families with Explicit Difficulty Control.}
In the current benchmark, explicit difficulty control is used for \emph{slider alignment}, \emph{jigsaw alignment}, \emph{missing-patch selection}, \emph{tile restoration}, and \emph{category-guided image selection}. Other task families are evaluated only under the default \texttt{normal} setting.

\paragraph{Slider Alignment.}
For \emph{slider alignment}, difficulty is controlled by both geometric precision and manipulation ease. In \texttt{easy}, the slider handle is enlarged and the positional tolerance is relaxed, which makes the target easier to align. In \texttt{hard}, the slider handle is reduced and the acceptance tolerance is tightened, increasing the precision required for success. 

\paragraph{Jigsaw Alignment.}
For \emph{jigsaw alignment}, difficulty is controlled through a combination of placement ambiguity and tolerance. The benchmark estimates the local visual contrast around candidate gap regions and uses this signal to choose where the missing piece is placed. In \texttt{easy}, the target gap is sampled from higher-contrast regions and the effective tolerance is relaxed; in \texttt{hard}, it is sampled from lower-contrast regions and the tolerance is tightened. As a result, harder instances place the missing piece in visually less distinctive regions while simultaneously demanding more precise alignment.

\paragraph{Missing-Patch Selection.}
For \emph{missing-patch selection}, difficulty is controlled by candidate ambiguity. In \texttt{easy}, the number of candidate patches is reduced. In \texttt{hard}, the number of candidates is increased and distractor candidates are additionally blurred, making local patch discrimination less reliable from coarse texture alone. \texttt{Normal} uses the default candidate count without the extra hard-mode distractor treatment.

\paragraph{Tile Restoration.}
For \emph{tile restoration}, difficulty is controlled by the number of swaps needed to restore the image. In the current static implementation, \texttt{hard} increases the scramble from one swapped pair to two swapped pairs, which expands the search space of possible corrections.

\paragraph{Category-Guided Image Selection.}
For \emph{category-guided image selection}, difficulty is controlled jointly by grid size, the number of correct targets, and the visual specificity of the image query. In \texttt{easy}, the task uses a smaller grid and exactly one correct image, and target retrieval favors close-up, visually salient examples. In \texttt{hard}, the grid becomes larger, the number of correct targets increases, and image retrieval shifts toward more scene-level or distant examples, which makes the target category visually less canonical. This design increases both perceptual ambiguity and the combinatorial burden of selecting all valid tiles.

\subsection{Dynamic Interaction Validation}
\label{app:dynamic}

Dynamic validation is implemented as a task-aware interaction-validation layer on top of semantic correctness. In dynamic settings, an agent must not only produce the correct answer, but also provide sufficient interaction evidence through a structured submission payload containing the final answer and the recorded telemetry stream.

\paragraph{Dynamic Validation Scope.}
Dynamic validation is enabled only for task families where correctness alone is insufficient to characterize a valid interaction and where the benchmark can define a stable family-level trace check. In the current study, dynamic support covers eight base families, while text transcription and logic-and-arithmetic interaction are treated as static-only families.

\paragraph{Mechanism Families.}
Rather than relying on one generic behavior score, \mybench \ organizes dynamic detection into several mechanism families according to the form of interaction evidence exposed by each task:
\begin{itemize}
    \item \textbf{Trajectory-continuity checks.}
    \item \textbf{Target-centered spatial-consistency checks.}
    \item \textbf{Repeated-wrong-loop checks.}
    \item \textbf{State-legality checks.}
\end{itemize}

\paragraph{Compositional Use.}
These mechanism families are not mutually exclusive. A single task may combine, for example, repeated-loop detection with state-legality constraints, or trajectory continuity with other task-specific safeguards. The grouping below therefore reflects recurring validation patterns in the implementation rather than a strict partition of task families.

\paragraph{Trajectory-Continuity Checks.}
This family is used for drag-based tasks such as \emph{slider alignment} and \emph{jigsaw alignment}. The benchmark records drag events such as \texttt{drag\_start}, \texttt{drag\_move}, and \texttt{drag\_end}, and then verifies whether a coherent drag chain actually occurred. The current implementation checks three properties in particular: whether drag interaction is present at all, whether the drag exhibits sufficient net displacement, and whether the motion avoids teleport-like jumps. This family is the clearest expression of continuous interaction validation in the benchmark.

\paragraph{Target-Centered Spatial-Consistency Checks.}
This family is used mainly for point-selection tasks such as \emph{natural-image sequence selection} and \emph{icon sequence selection}. Here the goal is not only to verify that clicking occurred, but to detect overly templated spatial behavior. In the current web implementation, the benchmark records \texttt{target\_click} events and compares click positions relative to the true target centers. It then checks whether the resulting offset pattern is implausibly consistent across multiple clicks. This design is more faithful than earlier size-normalized heuristics because it operates directly on target-centered relative offsets.

In addition to offset-consistency checks, these tasks may still require a minimally plausible click trace. However, the distinctive signal of this family is spatial regularity relative to actual targets rather than generic event presence alone.

\paragraph{Repeated-Wrong-Loop Checks.}
This family captures pathological retry behavior in selection- or control-driven tasks. The benchmark records temporally ordered interaction histories and checks whether the same erroneous operation pattern is repeated multiple times. Such repeated cycles are especially informative in agent evaluation because they reveal failure modes that are not visible from final-answer correctness alone.

\emph{Repeated wrong loops} are used primarily in selection- or enumeration-style tasks. For \emph{category-guided image selection}, the benchmark records \texttt{tile\_toggle} and \texttt{selection\_change} events, checks that real selection activity occurred, verifies that enough grid cells were actually touched, and detects whether the same wrong selection pattern is repeated multiple times. For \emph{missing-patch selection}, the benchmark records \texttt{candidate\_click} and \texttt{candidate\_select} events and detects repeated wrong candidate patterns.

\paragraph{State-Legality Checks.}
This family is used in board- and puzzle-like tasks where the intermediate state itself carries meaning. For \emph{tile restoration}, the benchmark records events such as \texttt{tile\_drag\_start}, \texttt{tile\_drag\_enter}, \texttt{tile\_drop}, \texttt{tile\_tap}, \texttt{swap\_commit}, and \texttt{initial\_order}. It checks whether each committed swap changed only a legal number of positions. For \emph{board reconfiguration}, the benchmark records \texttt{cell\_click}, \texttt{piece\_drag\_start}, \texttt{piece\_drop}, and \texttt{move\_commit}, then checks whether each state change remained within legal single-step bounds and whether repeated move patterns were detected.

\paragraph{Auxiliary Evaluation Safeguards.}
Some task families additionally include lightweight integrity checks over submitted payloads and simulator-recorded terminal states. We treat these as auxiliary benchmark-side safeguards for agent evaluation and failure analysis, rather than as primary dynamic detection mechanisms, because they help diagnose trace/payload inconsistencies but do not correspond to realistic production-side defenses.

These safeguards appear in several forms. For \emph{missing-patch selection}, the submitted answer can be compared against the last recorded candidate selection. For \emph{tile restoration} and \emph{board reconfiguration}, auxiliary checks may compare the final submission against the last recorded swap or move state. We retain these checks because they are useful for agent-side diagnosis and failure attribution, but we do not count them as part of the core dynamic mechanism taxonomy.

\paragraph{Representative Dynamic Signals.}
The benchmark records family-specific telemetry rather than relying on a single generic trajectory metric. Across the mechanism families above, representative signals include:
\begin{itemize}
    \item whether a coherent drag chain with sufficient displacement was recorded,
    \item whether click offsets are implausibly consistent relative to target centers,
    \item whether repeated wrong loops appear in selection or control history,
    \item and whether state changes exceed legal single-step bounds.
\end{itemize}

\paragraph{Consistency Checks.}
These interaction signals are converted into a small set of validation checks that are shared conceptually across task families, even though their concrete thresholds and rules are task-specific:
\begin{itemize}
    \item \textbf{Evidence-presence checks:} whether the submission contains the minimum telemetry required for validation.
    \item \textbf{Trajectory-continuity checks:} whether drag-based behavior exhibits real movement and avoids teleport-like jumps.
    \item \textbf{Spatial-consistency checks:} whether target-centered click offsets are implausibly regular.
    \item \textbf{Loop checks:} whether the same wrong interaction pattern is repeated multiple times.
    \item \textbf{State-legality checks:} whether intermediate state transitions remain within legal single-step bounds.
\end{itemize}

\paragraph{Dynamic Failure Taxonomy.}
At a high level, dynamic validation can reject an attempt for at least five reasons:
\begin{itemize}
    \item \textbf{Missing interaction evidence:} the submission lacks sufficient telemetry for dynamic verification.
    \item \textbf{Trajectory-continuity violation:} drag-based behavior lacks coherent movement evidence or contains teleport-like jumps.
    \item \textbf{Target-centered spatial anomaly:} click behavior is implausibly regular relative to target-centered offsets.
    \item \textbf{Repeated wrong loop:} the same erroneous interaction pattern is repeated multiple times.
    \item \textbf{Illegal state transition:} board or puzzle updates exceed legal single-step changes.
\end{itemize}

\paragraph{Supported Families and Primary Mechanisms.}
For transparency, we group dynamic-support families by their dominant recurring dynamic mechanisms:
\begin{itemize}
    \item \textbf{Trajectory continuity:} slider alignment and jigsaw alignment.
    \item \textbf{Target-centered spatial consistency:} natural-image sequence selection and icon sequence selection.
    \item \textbf{Repeated wrong loops:} category-guided image selection and missing-patch selection.
    
    \item \textbf{State legality:} tile restoration and board reconfiguration.
\end{itemize}

\paragraph{Per-Family Dynamic Validation Rules.}
Beyond the abstract mechanism categories above, each dynamic CAPTCHA variant instantiates a concrete set of validation rules. We summarize the effective validation logic of each family below, focusing on what is checked rather than on page-side implementation details.

\begin{itemize}
    \item \textbf{Slider alignment:} requires a real drag chain with both start and end events, sufficient net displacement, and no teleport-like jump in the recorded trajectory. A trial fails if the slider appears to move too little or if the path is implausibly discontinuous.

    \item \textbf{Jigsaw alignment:} uses the same trajectory-continuity principle as slider alignment. The dynamic check requires a genuine drag trajectory for the puzzle piece, sufficient movement magnitude, and the absence of jump-like discontinuities.

    \item \textbf{Natural-image sequence selection:} requires enough recorded clicks to cover the expected sequence, rejects traces whose click positions collapse to too few distinct locations, and then checks whether click offsets relative to the true target centers are implausibly consistent across multiple selections.

    \item \textbf{Icon sequence selection:} follows the same target-centered spatial-consistency logic as natural-image sequence selection. The dynamic check requires enough icon clicks, sufficient positional diversity, and rejects unnaturally templated relative offsets around icon centers.

    \item \textbf{Category-guided image selection:} requires enough recorded tile toggles to support the submitted selection, checks that the interaction trail touches enough distinct grid cells, and then detects repeated wrong selection loops by testing whether the same erroneous toggle/selection pattern is repeated multiple times.

    \item \textbf{Missing-patch selection:} primarily uses repeated-wrong-selection detection. The benchmark checks whether a real candidate selection was recorded and whether the same wrong candidate pattern is repeatedly retried. In addition, auxiliary benchmark-side safeguards may compare the submitted answer against the final recorded candidate selection.

    \item \textbf{Tile restoration:} requires that a real swap interaction be observed through either a tap-based or drag-based flow, and then checks state legality by verifying that each committed swap changes only a legal number of positions. Auxiliary benchmark-side safeguards may additionally compare the submitted answer against the last recorded swap state.

    \item \textbf{Board reconfiguration:} requires that a real move be observed through click-flow or drag-flow, checks that each committed board update remains within legal single-step bounds, and detects repeated move patterns when the same erroneous sequence recurs. Auxiliary safeguards may additionally compare the submitted answer against the final recorded move state.
\end{itemize}

\paragraph{Design Rationale.}
This organization keeps the dynamic layer analyzable without reducing it to a brittle one-size-fits-all score. It also matches the actual implementation more closely: drag tasks are characterized by trajectory continuity, point-selection tasks by target-centered spatial regularity, some tasks by retry loops, and others by state-legality violations. Auxiliary integrity checks remain available for benchmark-side diagnosis and agent analysis, but they are intentionally separated from the core dynamic mechanism taxonomy because they do not correspond to realistic deployment-side defenses. As a result, dynamic validation can be analyzed not only by whether an agent passes or fails, but also by which type of anti-automation mechanism it consistently struggles with.

\subsection{Benchmark Statistics and Support Matrix}
\label{app:benchmark-stats}
Table~\ref{tab:benchmark-support-matrix} summarizes the benchmark support matrix used in the main experiments. The benchmark contains ten base task families. All families support the clean static setting and the webpage-distraction setting. Five families additionally support explicit hard variants, and eight families support dynamic interaction validation. Counting each supported family--setting cell once yields 33 evaluated settings in total: ten clean static settings, ten distraction settings, five hard settings, and eight dynamic settings.

\begin{table*}[t]
\centering
\small
\setlength{\tabcolsep}{5pt}
\caption{Benchmark support matrix across task families and realism settings. ``Yes'' indicates that the corresponding family is evaluated under that setting in the main study.}
\label{tab:benchmark-support-matrix}
\resizebox{\textwidth}{!}{
\begin{tabular}{lcccc}
\hline
Task family & Clean static & Webpage distraction & Hard variant & Dynamic validation \\
\hline
Text transcription & Yes & Yes & -- & -- \\
Slider alignment & Yes & Yes & Yes & Yes \\
Natural-image sequence selection & Yes & Yes & -- & Yes \\
Jigsaw alignment & Yes & Yes & Yes & Yes \\
Logic-and-arithmetic interaction & Yes & Yes & -- & -- \\
Category-guided image selection & Yes & Yes & Yes & Yes \\
Missing-patch selection & Yes & Yes & Yes & Yes \\
Icon sequence selection & Yes & Yes & -- & Yes \\
Tile restoration & Yes & Yes & Yes & Yes \\
Board reconfiguration & Yes & Yes & -- & Yes \\
\hline
Total supported settings & 10 & 10 & 5 & 8 \\
\hline
\end{tabular}
}
\end{table*}

The support matrix reflects the benchmark design rather than an arbitrary omission of results. Text transcription is retained as a static recognition baseline and is not assigned a dynamic variant in the main definition. Logic-and-arithmetic interaction is evaluated under clean and distracted webpage conditions, but its heterogeneous subtype structure makes a single hard-setting control or family-level dynamic mechanism less directly comparable than the controls used for alignment, selection, patch, restoration, and board-manipulation tasks. Board reconfiguration supports dynamic validation through state-legality checks, but its hard-setting extension is not included in the main study to keep the difficulty-controlled comparison aligned across families with explicit and directly comparable hardness parameters.

\section{Additional Experimental Details and Full Results}
\label{app:exp-details}

This appendix provides supplementary material for Section~\ref{sec:experiments}. It collects experimental configuration details and extended qualitative analyses that clarify the failure modes behind those results. Unless otherwise specified, all results in this appendix follow the same evaluation protocol defined in Section~\ref{subsec:exp-setup}.

\subsection{Detailed Experimental Setup}
\label{app:exp-setup}

This subsection summarizes the evaluation configuration used in our experiments, including:
\begin{itemize}
    \item the evaluated model and agent list,
    \item the step budget, timeout, and attempt limits where applicable,
    \item the number of sampled instances per configuration where reported,
    \item and any setting-specific implementation notes needed for reproducibility.
\end{itemize}

All experiments are conducted with the same closed-loop evaluation pipeline. In each batch, one or more agents are evaluated on a selected set of CAPTCHA families under a fixed configuration specifying the number of sampled instances per family, the maximum number of interaction steps, the maximum number of attempts, the per-sample timeout, and the realism controls used in that batch, including difficulty and distraction level when applicable.

The experimental study is organized into several batches rather than a single monolithic sweep. The clean static setting evaluates the ten base task families under default task configuration and distraction level~0. For families with explicit difficulty control, we additionally evaluate \texttt{easy}, \texttt{normal}, and \texttt{hard} variants. For webpage-clutter analysis, we evaluate increasing distraction levels. Dynamic experiments are run on the corresponding dynamic task variants and use the same simulator setting, except that a semantically correct answer must additionally satisfy the family-specific trace-conditioned dynamic validation rules defined in Section~\ref{subsec:realism-axes}.

Across all batches, the evaluation protocol is consistent: an agent interacts with the rendered verification page in a closed loop until it submits an answer, reaches the step limit, exhausts the allowed attempts, or times out. In the current experiments, the per-sample timeout is set to 1200 seconds, while the interaction budget and the number of sampled instances are fixed within each batch. This batch-wise reporting is important because the present study was collected incrementally across static, difficulty-controlled, distraction-controlled, and dynamic settings.

For each evaluated sample, the evaluation pipeline records both aggregate outputs and detailed artifacts. We retain the final result record together with interaction traces and supporting visual evidence such as screenshots when available. 
% These records are used not only for success-rate reporting, but also for the extended qualitative analysis in Appendix~\ref{app:failure-cases}.

Unless otherwise specified, all reported metrics are computed at the instance level. A static trial is counted as successful if the final submitted answer matches the ground-truth solution. A dynamic trial must first satisfy static correctness and then pass the corresponding trace-conditioned dynamic validation rule over simulator-observable interaction traces and committed page states. This shared setup ensures that comparisons across base, distraction, difficulty, and dynamic settings remain directly aligned.

\subsection{Full Static Results Analysis}
\label{app:full-static}

Table~\ref{tab:static-base-results} reports the complete static results under clean webpages and default task configurations. These results complement the discussion in Section~\ref{subsec:static-base-results} by exposing the full cross-model and cross-family structure of performance. Two high-level patterns are immediately visible. First, static performance varies substantially across models: Claude-Opus-4.6 is the strongest model in this setting, while several other agents remain effective only on a subset of families. Second, performance is highly non-uniform across task families, indicating that even without distraction, elevated difficulty, or dynamic validation, CAPTCHA solving still requires a composition of heterogeneous capabilities rather than a single generic perceptual skill.

At the family level, text transcription is clearly the easiest task in the clean static setting, with a mean success rate of 84.8 across evaluated models. Several families involving direct visual localization or single-step recognition are also moderately solvable on average, including jigsaw alignment, slider alignment, and logic-and-arithmetic interaction. In contrast, icon-sequence selection and tile restoration are the least solved families, with mean success rates of 21.0 and 25.3. This ranking is consistent with the intended capability structure of the benchmark: the easier families are those where recognition dominates, whereas the harder families require stronger sequential control, fine-grained spatial grounding, or persistent manipulation of an evolving interface state.

The model-level profiles are similarly heterogeneous. Claude-Opus-4.6 achieves the strongest overall static performance, reaching near-perfect success on most task families and maintaining a macro-average of 90.0. Its remaining errors are concentrated on ordered icon selection and board reconfiguration, suggesting that even the strongest model is not fully saturated on tasks requiring ordered target interaction or stateful layout reasoning. Gemini-3.1-Pro is the most balanced among the remaining models, with consistently competitive performance across almost all families. GPT-5.4 exhibits a more polarized profile: it is nearly saturated on text transcription, jigsaw alignment, category-guided image selection, and missing-patch selection, but drops sharply on icon-sequence selection and tile restoration. The lower-performing group, including Claude-Sonnet-4.6, GLM-5V, MiniMax-M2.7, and Qwen-Max, further confirms that strong performance on one or two families does not translate into robust end-to-end capability across the benchmark.

\begin{table}[t]
\centering
\small
\caption{Family-wise summary of static results under clean webpages and default task configurations. ``Mean'' denotes the average success rate (\%) across all evaluated models in the current static sweep. ``Best model'' reports the strongest model on that family under the same setting.}
\label{tab:app-static-family-summary}
\begin{tabular}{lcc}
\hline
Family & Mean & Best model \\
\hline
Text transcription & 84.8 & GPT-5.4 / Gemini-3.1-Pro / Claude-Sonnet-4.6 / Claude-Opus-4.6 \\
Slider alignment & 54.0 & Claude-Opus-4.6 \\
Image-sequence selection & 43.0 & Claude-Opus-4.6 \\
Jigsaw alignment & 55.5 & Claude-Opus-4.6 \\
Logic-arithmetic interaction & 53.3 & Claude-Opus-4.6 \\
Category-guided image selection & 47.3 & GPT-5.4 / Claude-Opus-4.6 \\
Missing-patch selection & 44.5 & Claude-Opus-4.6 \\
Icon-sequence selection & 21.0 & Claude-Opus-4.6 \\
Tile restoration & 25.3 & Claude-Opus-4.6 \\
Board reconfiguration & 38.0 & Gemini-3.1-Pro \\
\hline
\end{tabular}
\end{table}

\subsubsection{Static Failure Cases}
\label{app:static-failure-cases}

We further inspect representative failed traces from the clean static setting. The taxonomy below organizes the observed errors into eight capability-aligned categories. Each category isolates a distinct bottleneck in the static setting, from low-level perception to higher-level state management and recovery.

\paragraph{Perceptual decoding errors.}
These failures correspond to weak fine-grained visual recognition and unstable OCR. The agent misreads distorted text, confuses visually similar characters, or fails to distinguish subtle local image details before any complex interaction is required.
% Case placeholder:
\begin{figure}[t]
\centering
\includegraphics[width=\linewidth]{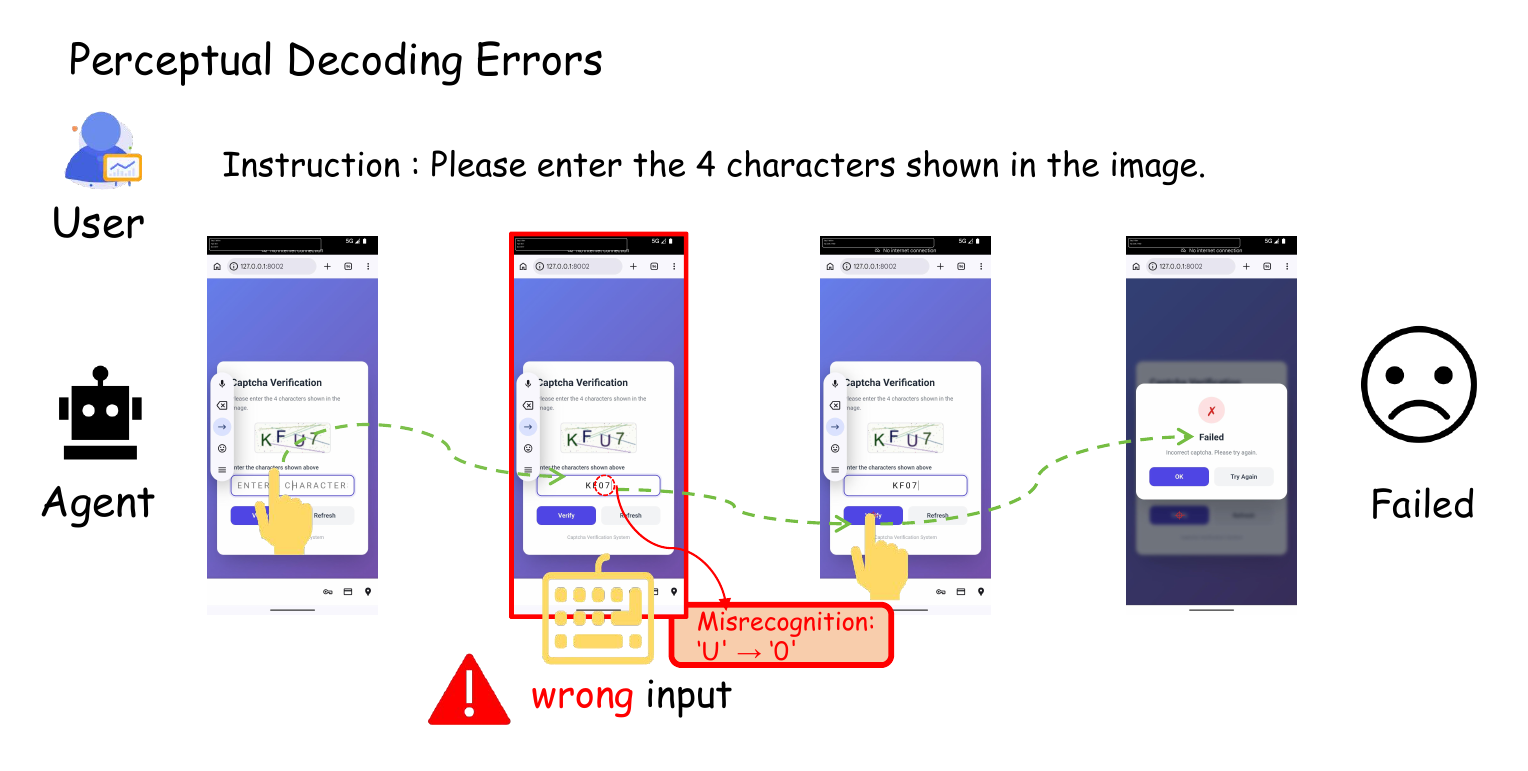}
\caption{Representative static failure case for perceptual decoding errors.}
\label{fig:static-perceptual-decoding-case}
\end{figure}

\paragraph{Target localization and candidate filtering errors.}
These failures correspond to inaccurate visual localization and insufficient robustness in target screening. The agent may understand the requested target category or sequence but fail to identify the correct candidate, miss valid targets, or over-select visually similar distractors.
% Case placeholder:
\begin{figure}[t]
\centering
\includegraphics[width=\linewidth]{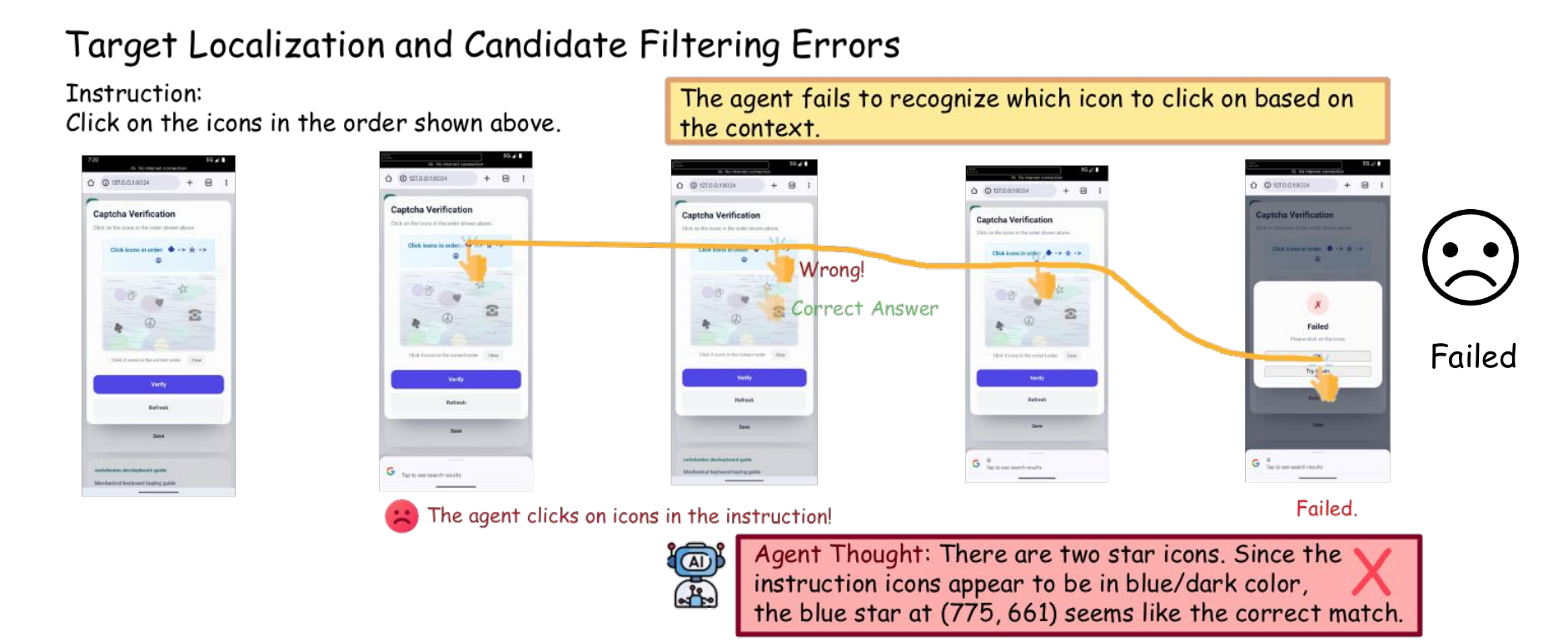}
\caption{Representative static failure case for target localization and candidate filtering errors.}
\label{fig:static-target-filtering-case}
\end{figure}

\paragraph{Spatial grounding and action mapping errors.}
These failures correspond to weak visual-to-action spatial mapping. The agent often has the right semantic target in mind, but maps that target to an incorrect click or drag coordinate, producing the common ``recognize correctly but act at the wrong location'' pattern.
% Case placeholder:
\begin{figure}[t]
\centering
\includegraphics[width=\linewidth]{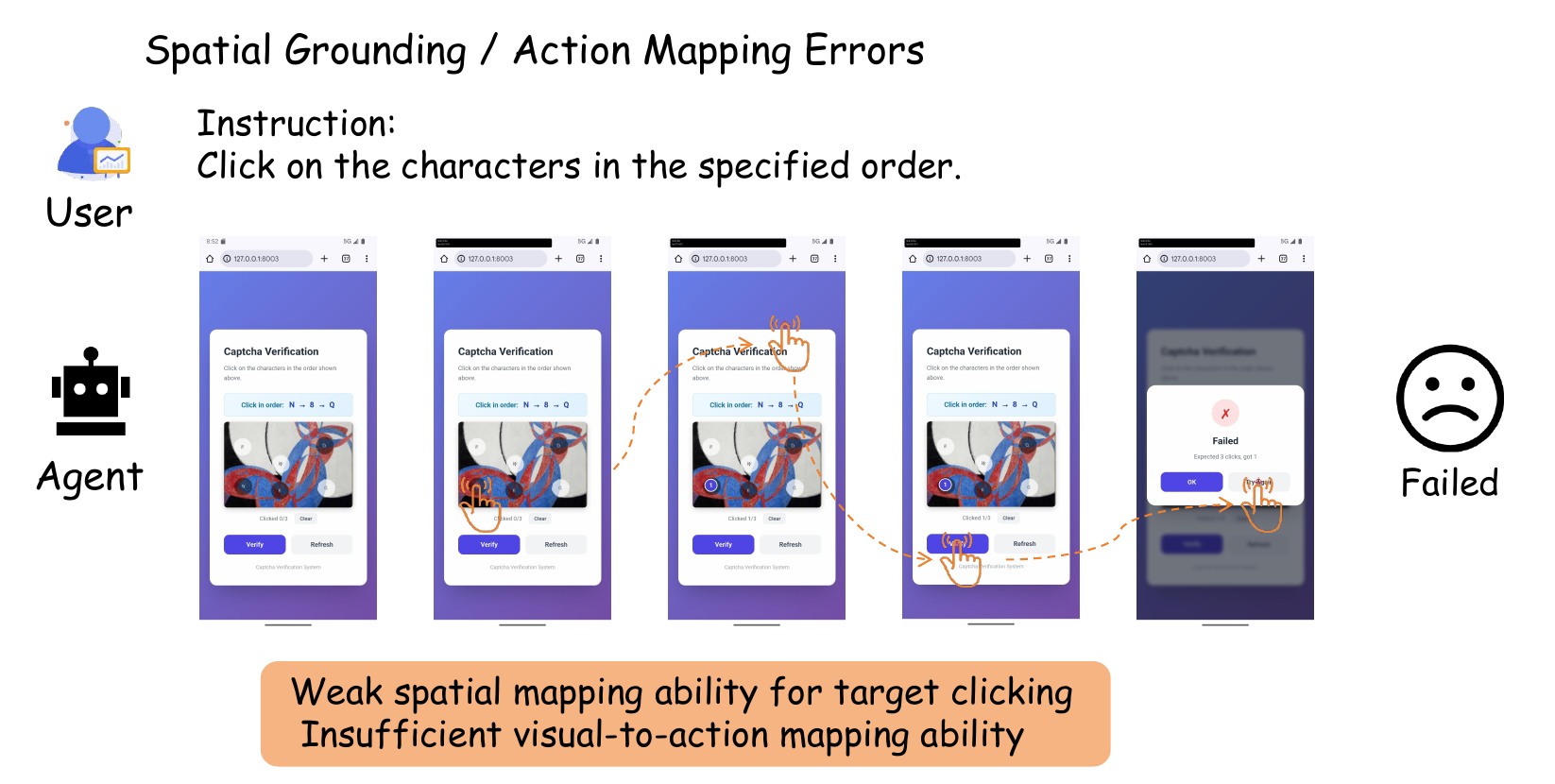}
\caption{Representative static failure case for spatial grounding and action mapping errors.}
\label{fig:static-spatial-grounding-case}
\end{figure}

\paragraph{Geometric calibration failures.}
These failures correspond to insufficient precision in geometric estimation and fine-grained control. Slider and jigsaw-alignment tasks expose this bottleneck most directly: agents often understand the intended operation, but choose an inaccurate drag distance, stop short of the target, overshoot it, or repeat a poorly calibrated movement.
% Case placeholder:
\begin{figure}[t]
\centering
\includegraphics[width=\linewidth]{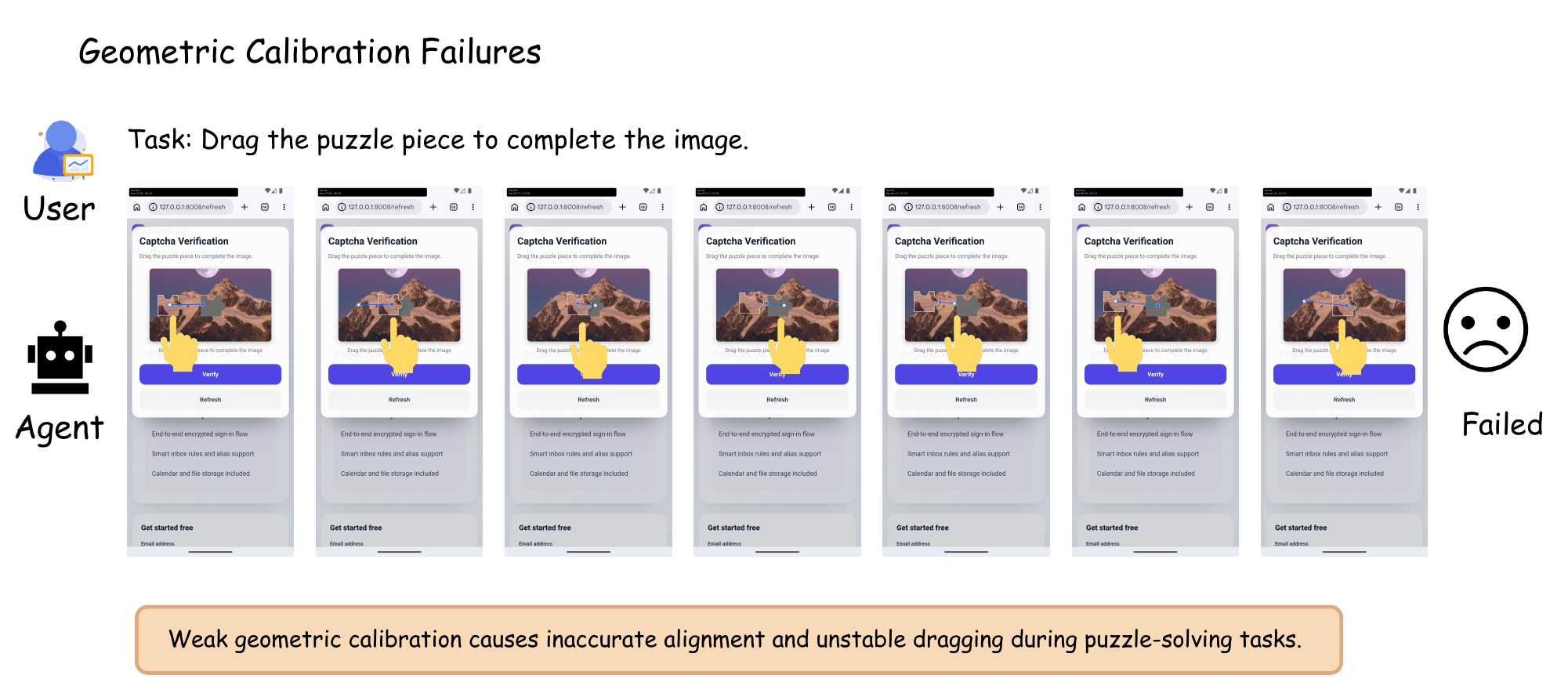}
\caption{Representative static failure case for geometric calibration failures.}
\label{fig:static-geometric-calibration-case}
\end{figure}

\paragraph{Visual structure reconstruction failures.}
These failures correspond to weak visual structure understanding and limited global-consistency reasoning. Tile restoration, board reconfiguration, missing-patch selection, and puzzle-like tasks require the agent to infer an underlying visual or spatial structure; failed traces often show uncertainty about which element is misplaced or how local changes affect the global configuration.
% Case placeholder:
\begin{figure}[t]
\centering
\includegraphics[width=\linewidth]{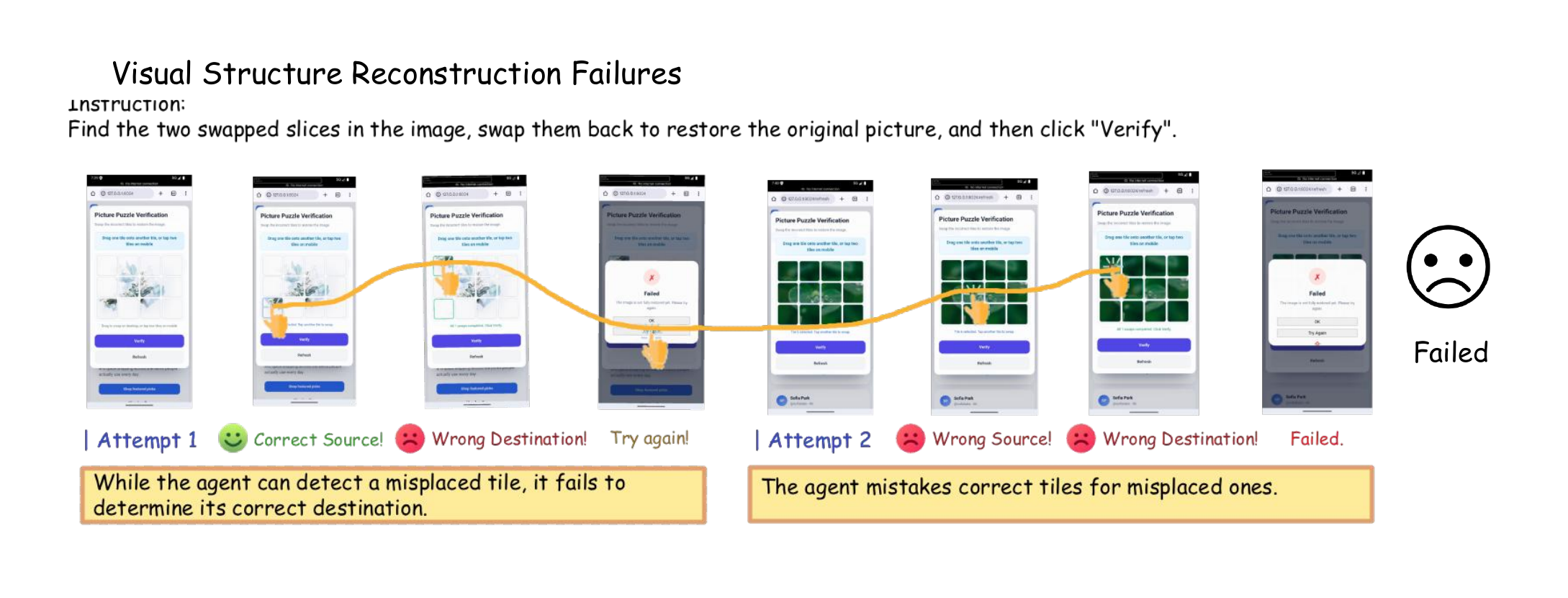}
\caption{Representative static failure case for visual structure reconstruction failures.}
\label{fig:static-structure-reconstruction-case}
\end{figure}

\paragraph{UI affordance and interactive-region misunderstanding.}
These failures correspond to insufficient UI-structure understanding and weak recognition of interactive regions. The agent may confuse the CAPTCHA region with surrounding controls, misidentify input fields or buttons, or fail to determine which visible elements are actually actionable.
% Case placeholder:
\begin{figure}[t]
\centering
\includegraphics[width=\linewidth]{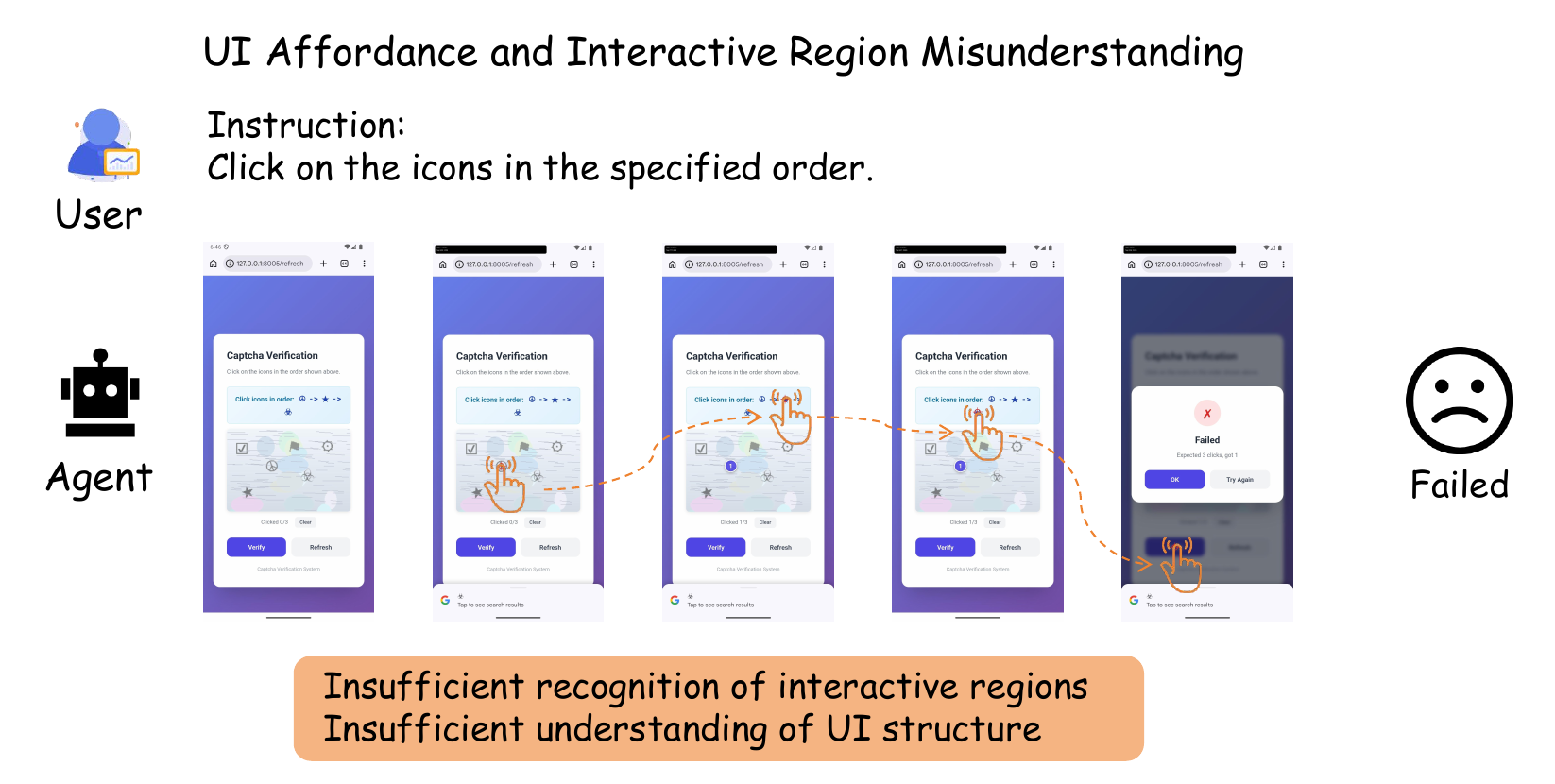}
\caption{Representative static failure case for UI affordance and interactive-region misunderstanding.}
\label{fig:static-ui-affordance-case}
\end{figure}

\paragraph{State tracking and completion-judgment failures.}
These failures correspond to weak state management, context isolation, and task-completion judgment. In multi-step tasks, the agent may lose track of which targets have already been selected, repeat completed steps, submit too early, or continue interacting after the task state has already changed.
% Case placeholder:
\begin{figure}[t]
\centering
\includegraphics[width=\linewidth]{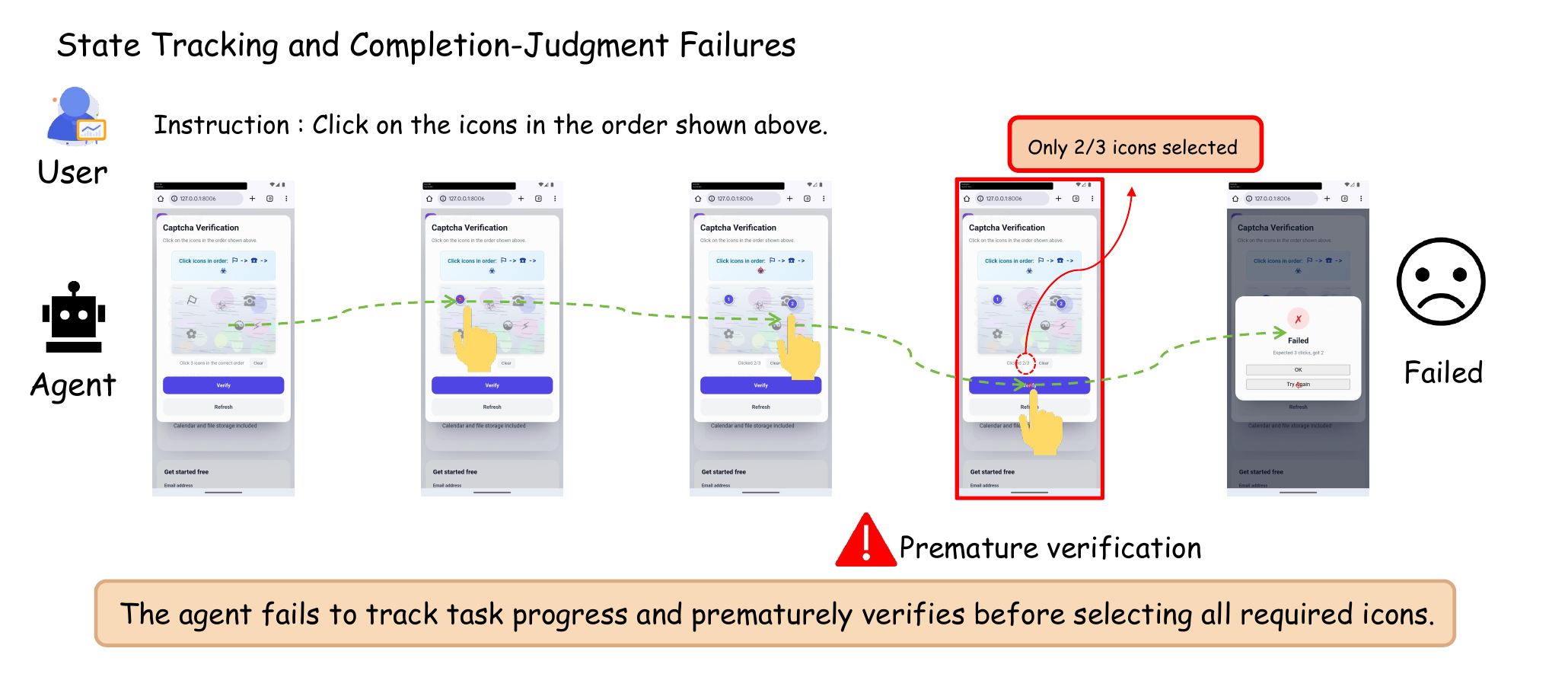}
\caption{Representative static failure case for state tracking and completion-judgment failures.}
\label{fig:static-state-tracking-case}
\end{figure}

\paragraph{Recovery and self-correction failures.}
These failures correspond to weak correction, abnormal-interface handling, and strategy adaptation after feedback. Across several families, agents recognize that an attempt has failed but enter low-value retry loops, refresh repeatedly, resubmit stale answers, repeat old coordinates, or emit malformed actions instead of adapting their next step.

\subsection{Additional Failure Modes under Distraction}
\label{app:full-distraction}

\subsubsection{Distraction Result Interpretation}
\label{app:distraction-result-tables}

The complete distraction results are reported in Table~\ref{tab:distraction-results}. Relative to the clean static setting, webpage distraction does not introduce a new CAPTCHA operation. Instead, it adds page-level context that makes the agent first identify the true verification region and suppress irrelevant surrounding content. We therefore focus only on failure modes that are newly exposed by the distracted webpage setting, rather than repeating the static taxonomy.

\subsubsection{New Distraction Failure Modes}
\label{app:distraction-failure-cases}

\paragraph{Task-region localization and distraction suppression failures.}
Unlike the clean setting, the distracted setting requires the agent to separate the active CAPTCHA card from ordinary webpage content, background text, decorative elements, and decoy controls. Failed traces often show that the agent either acts on surrounding page elements or spends excessive steps re-reading the page without committing to the true verification region. This failure is distinct from static target localization because the ambiguity occurs at the page level before the CAPTCHA content itself is solved.

\paragraph{Layout-shift spatial generalization failures.}
Distraction also changes the position, scale, and surrounding context of the verification widget. Some agents appear to rely on coordinate priors learned from clean layouts, repeatedly clicking familiar relative positions even when the target has shifted or the card is embedded differently. This newly exposes weak spatial generalization across page layouts: the agent may still know the CAPTCHA task type, but fail to remap its actions after the interface is placed in a more realistic webpage context.

\subsection{Additional Failure Modes under Hard Variants}
\label{app:full-difficulty}

\subsubsection{Difficulty Result Interpretation}
\label{app:difficulty-result-tables}

The complete hard-setting results are reported in Table~\ref{tab:hard-results}. Hard variants mostly amplify the static failures described above: perception errors become more frequent, action calibration has less room for error, and state reconstruction becomes harder. The additional analysis below therefore focuses on the failure modes that become newly salient only when the task is made intrinsically harder.

\subsubsection{New or Amplified Hard-Setting Failure Modes}
\label{app:difficulty-failure-cases}

\paragraph{High-similarity discrimination failures.}
Hard image-selection and missing-patch tasks increase candidate similarity or visual ambiguity. This exposes a sharper form of fine-grained discrimination failure: agents can often identify obvious positives, but fail on borderline candidates, visually similar distractors, or local patches that require comparing subtle texture and shape consistency.

\paragraph{Strict-tolerance calibration failures.}
Hard slider and jigsaw-alignment tasks reduce the acceptable error tolerance. As a result, actions that are semantically correct but slightly imprecise no longer pass. The newly salient issue is not merely that the agent maps a target to the wrong coordinate, but that it cannot reliably produce sufficiently precise movements when the success region becomes narrow.

\paragraph{Complex state-search failures.}
Hard restoration tasks require more complex search over possible intermediate states, such as identifying multiple misplaced tiles instead of a single obvious swap. This setting newly emphasizes the agent's inability to plan and verify a multi-step correction sequence under a larger state space, even when it can describe the overall puzzle structure.

\subsection{Dynamic Validation Analysis}
\label{app:full-dynamic}

\subsubsection{Dynamic Result Interpretation}
\label{app:dynamic-result-tables}

The dynamic results in Table~\ref{tab:dynamic-results} show that trace-conditioned validation exposes a process-level weakness that is not captured by static correctness. Across the eight dynamic-supported task families, Gemini-3.1-Pro obtains the highest average success rate (45.0), followed by GPT-5.4 (26.3) and Claude-Opus-4.6 (23.8). This ranking differs from the base static setting, where Claude-Opus-4.6 is the strongest overall model. The reversal suggests that dynamic validation is not merely a harder version of static recognition; it tests whether the model can produce an interaction trajectory that remains valid under task-specific evidence checks.

The strongest degradation appears on task families whose success depends on the relation between the final answer and the action trace. Slider alignment and jigsaw alignment require coherent drag evidence, ordered clicking tasks require target-centered spatial consistency, and restoration-style tasks require legal state transitions. In these families, an answer can be visually or semantically plausible while still failing dynamic validation because the observed trace does not support the submitted result.

\subsubsection{Mechanism-Level Dynamic Failures}
\label{app:dynamic-failure-cases}

Dynamic failures can be organized into three additional process-level categories that are distinct from the static failure taxonomy.

\paragraph{Insufficient interaction evidence.}
The agent may submit or reach a seemingly correct final answer without producing enough task-relevant evidence in the recorded trace. This occurs when required clicks, selections, drags, or state changes are missing or too sparse to support the final answer. Such cases are especially important because they would be counted as plausible successes under a purely static metric, but fail once the benchmark checks whether the answer is backed by observable interaction.

\paragraph{Trajectory-continuity and action-authenticity failures.}
Drag-based tasks such as slider alignment and jigsaw alignment require a coherent motion trace, not just a final coordinate. Dynamic validation therefore exposes failures where the recorded action sequence lacks a valid drag start--move--end chain, contains teleport-like jumps, or does not show enough continuous displacement. These failures explain why some models that understand the alignment target still obtain low dynamic success on alignment families.
\begin{figure}[t]
\centering
\includegraphics[width=\linewidth]{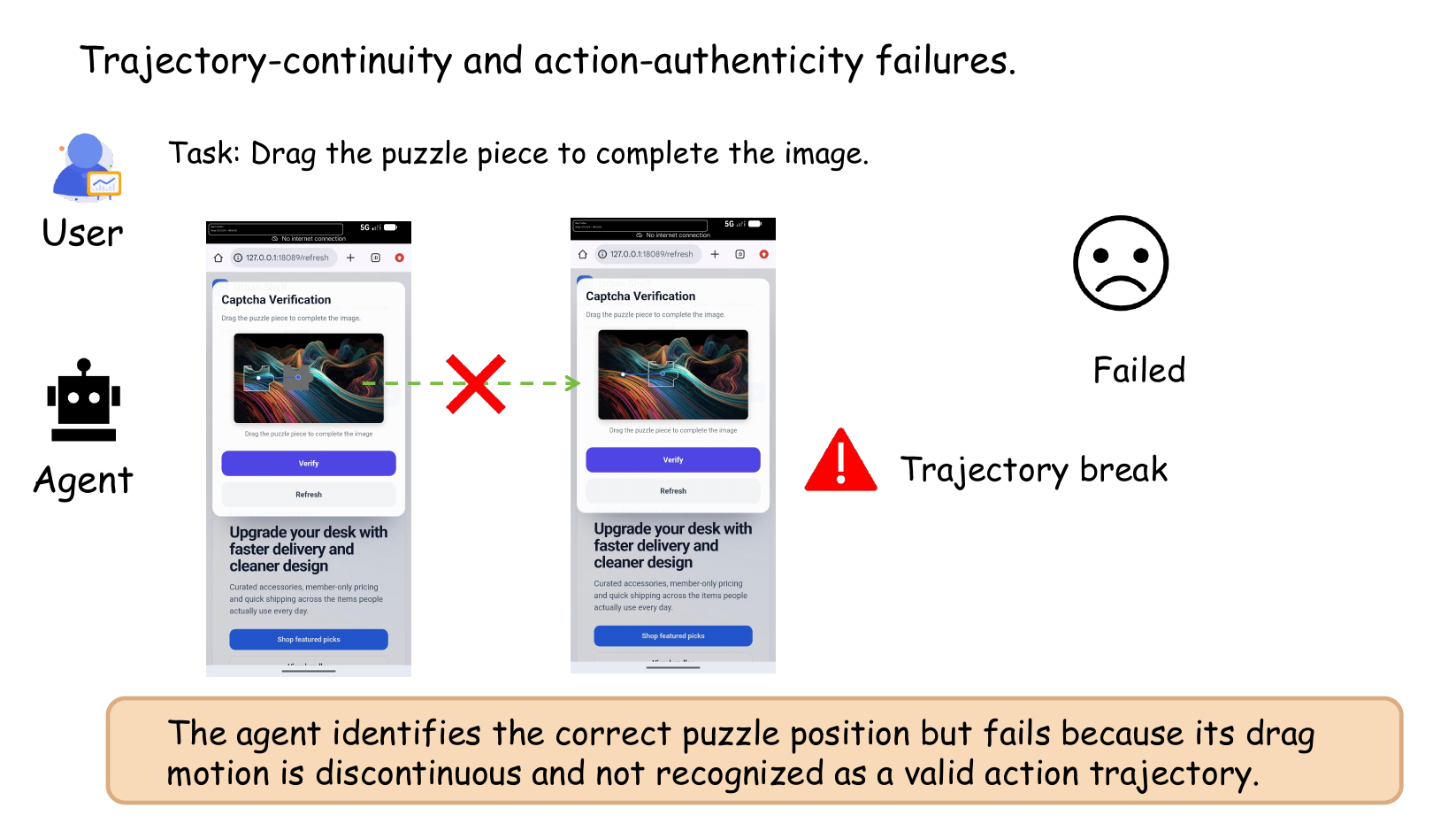}
\caption{Representative dynamic failure case for Trajectory-continuity and action-authenticity failures.}
\label{fig:dynamic-case}
\end{figure}

\paragraph{Interaction-history and terminal-state inconsistency.}
For ordered clicking, board reconfiguration, tile restoration, and selection-based tasks, the submitted answer must be consistent with the sequence of previous actions and the final page state. Dynamic failures arise when the action history contains repeated wrong clicks, illegal transitions, incomplete selections, or a final state that does not match the submitted result. This category captures cases where the model appears to know the answer but cannot maintain a valid process from initial observation to terminal submission.

\subsection{Responsible Release and Intended Use}
\label{app:responsible-release}

We position \mybench \ as a controlled simulator and evaluation harness for studying interactive verification, not as tooling to attack or bypass arbitrary live third-party verification systems. Release materials are intended to reproduce the benchmark episodes, metrics, and validation rules reported in this paper, together with documentation.

\end{document}